\documentclass[10pt,twocolumn,letterpaper]{article}

\usepackage[pagenumbers]{cvpr} 

\usepackage{graphicx}
\usepackage{amsmath}
\usepackage{amssymb}
\usepackage{booktabs}

\usepackage[pagebackref,breaklinks,colorlinks]{hyperref}

\usepackage[utf8]{inputenc} 
\usepackage[T1]{fontenc}    
\usepackage{hyperref}       
\usepackage{url}            
\usepackage{booktabs}       
\usepackage{amsfonts}       
\usepackage{nicefrac}       
\usepackage{microtype}      
\usepackage{xcolor,colortbl}          
\usepackage{multirow}
\usepackage{graphicx}
\usepackage{algorithm}
\usepackage{algorithmic}
\usepackage{wrapfig}
\usepackage{diagbox}
\usepackage{color}
\usepackage{amsmath}
\usepackage{amssymb}

\usepackage{subcaption}
\usepackage{amsthm}

\newtheorem{case}{Case}

\usepackage{xcolor}
\newcommand\blfootnote[1]{%
  \begingroup
  \renewcommand\thefootnote{}\footnote{#1}%
  \addtocounter{footnote}{-1}%
  \endgroup
}
\newcommand{\lv}[1]{{\color{black}{#1}}}
\newcommand{\lvv}[1]{{\color{black}{#1}}}

\usepackage[capitalize]{cleveref}
\crefname{section}{Sec.}{Secs.}
\Crefname{section}{Section}{Sections}
\Crefname{table}{Table}{Tables}
\crefname{table}{Tab.}{Tabs.}

\def\cvprPaperID{12384} 
\def\confName{CVPR}
\def\confYear{2023}

\begin{document}

\title{Improving Generalization with Domain Convex Game}

\author{
\textbf{Fangrui Lv}\textsuperscript{\rm 1}
~
\textbf{Jian Liang}
~
\textbf{Shuang Li}\textsuperscript{\rm 1,$*$}
~
\textbf{Jinming Zhang}\textsuperscript{\rm 1}
~
\textbf{Di Liu}
\\ [0.25cm]
\textsuperscript{\rm 1} Beijing Institute of Technology, China
\\[0.1cm]
{$^1$ \tt\small \{fangruilv,shuangli,jinming-zhang\}@bit.edu.cn}
~~
{\tt\small \{liangjianzb12,liudi010\}@gmail.com}
}

\maketitle

\begin{abstract}
Domain generalization (DG) tends to alleviate the poor generalization capability of deep neural networks by learning model with multiple source domains. 
A classical solution to DG is domain augmentation, the common belief of which is that diversifying source domains will be conducive to the out-of-distribution generalization. 
However, these claims are understood intuitively, rather than mathematically. 
Our explorations empirically reveal that the correlation between model generalization and the \lvv{diversity of domains} \lvv{may be} not strictly positive, which limits the effectiveness of domain augmentation. This work therefore aim to guarantee and further enhance the validity of this strand.
To this end, we propose a new perspective on DG that recasts it as a convex game between domains. We first encourage each diversified domain to enhance model generalization by elaborately designing a regularization term based on supermodularity. Meanwhile, a sample filter is constructed to eliminate low-quality samples, \lvv{thereby avoiding the impact of potentially harmful information}.
Our framework presents a new avenue for the formal analysis of DG, heuristic analysis and extensive experiments demonstrate the rationality and effectiveness.
\blfootnote{$*$ Corresponding author.}
\footnote{\quad Code is available at "https://github.com/BIT-DA/DCG".}

\vspace{-4mm}

\end{abstract}

\section{Introduction}
\label{sec:intro}
Owning extraordinary representation learning ability, deep neural networks (DNNs) have achieved remarkable success on a variety of tasks when the training and test data are drawn from the same distribution ~\cite{DeepLearning,resnet,DeepLearning2}. 
Whereas for out-of-distribution data, DNNs have demonstrated poor generalization capability since the i.i.d. assumption is violated, which is common in real-world conditions~\cite{RTN,MMAN,robustness}. 
To tackle this issue, domain generalization (DG) has become a propulsion technology, aiming to learn a robust model from multiple source domains so that can generalize well to any unseen target domains with different statistics~\cite{MetaReg,DICA,MMD-AAE,DBA}. 

Among extensive solutions to improve generalization, domain augmentation~\cite{AdvAug,CrossGrad,MixStyle,FACT} has been a classical and prevalent strategy, which focuses on exposing the model with more diverse domains via some augmentation techniques. A common belief is that generalizable models would become easier to learn when the training distributions become more diverse, which has been also emphasized by a recent work~\cite{DomainAug}. 
Notwithstanding the promising results shown by this strand of approaches, the claims above are vague and lack of theoretical justification, formal analyses of the relation between domain diversity and model generalization are sparse. 
Further, the transfer of knowledge may
even hurt the performance on target domains in some cases, which is referred to as negative transfer~\cite{TransferLearning, DistantTL}. Thus the relation of domain diversity and model generalization remains unclear. In light of these points, we begin by considering the question: \textbf{The stronger the domain diversity, will it certainly help to improve the model generalization capability?}

\begin{figure}[t]
    \centering
    \begin{subfigure}{0.493\linewidth}
      \includegraphics[width=1.0\linewidth]{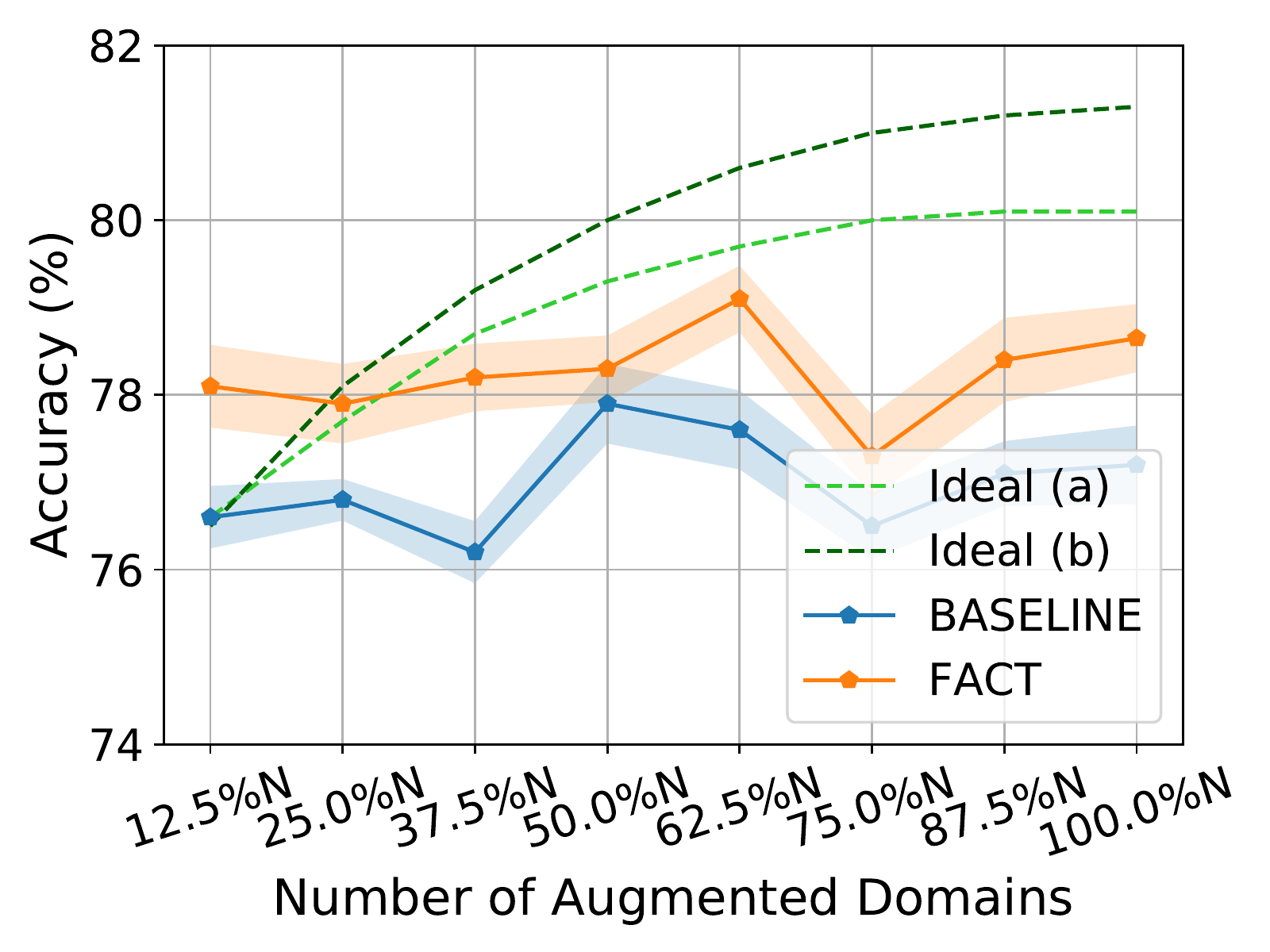}
      \caption{Cartoon.}
    \end{subfigure}
    \hfill
    \begin{subfigure}{0.493\linewidth}
    \includegraphics[width=1.0\linewidth]{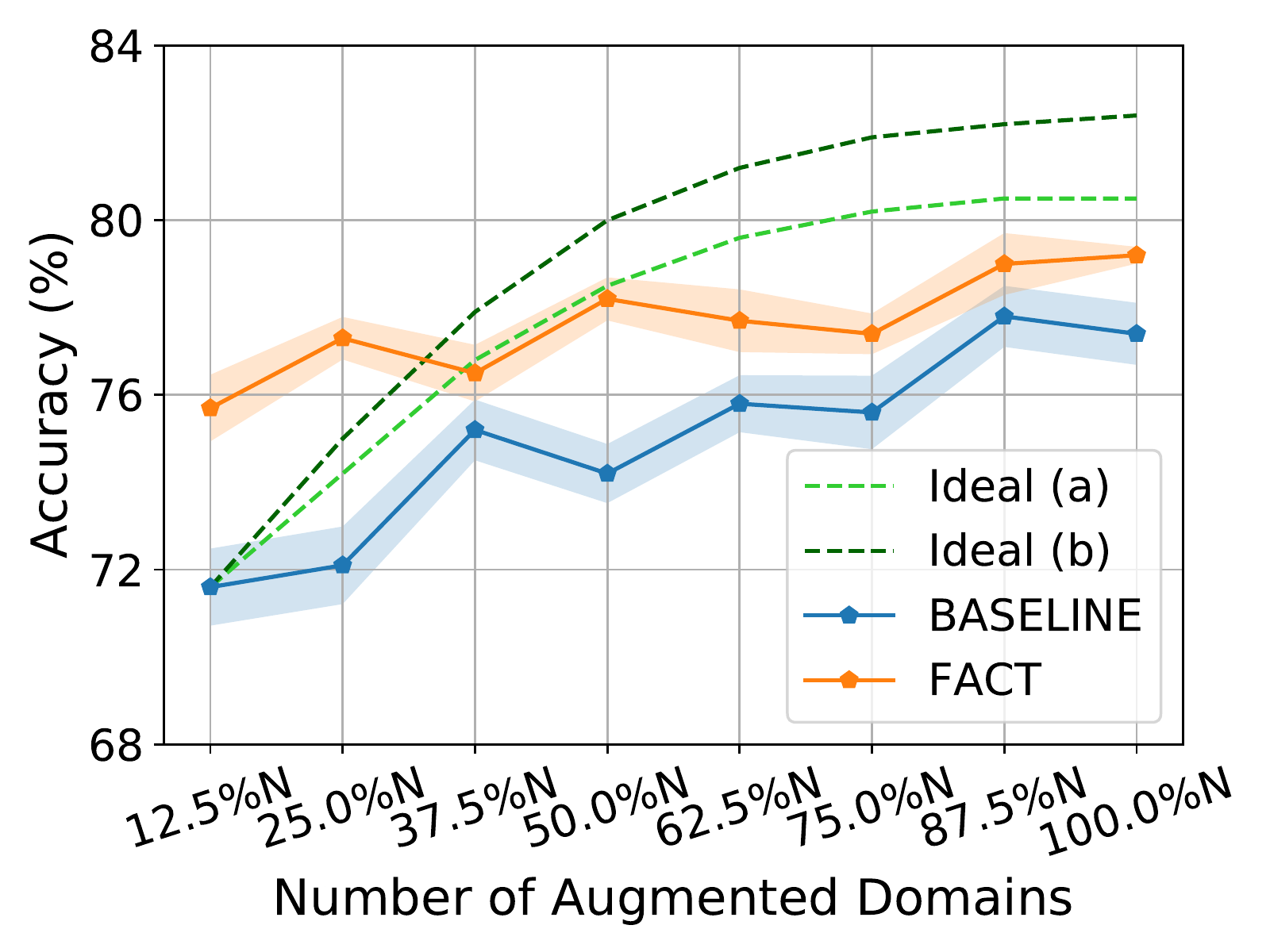}
      \caption{Sketch.}
    \end{subfigure}
    \caption{The relation between model generalization and domain diversity with Cartoon and Sketch on PACS dataset as the unseen target domain, respectively. \lv{$N$ is the maximum number of augmented domains. Note that the \textit{solid lines} denote the actual results of a BASELINE method that combines DeepAll with Fourier augmentation strategy and a SOTA domain augmentation method FACT, while the \textit{dash lines} represent the ideal relation in this work. }}
    \label{fig:motivation}
    \vspace{-8mm}
  \end{figure}

\lvv{To explore this issue, we first quantify domain diversity as the number of augmented domains. Then we conduct a brief experiment using Fourier augmentation strategy~\cite{FACT} as a classical and representative instance.}
The results presented in Fig~\ref{fig:motivation} show that with the increase of domain diversity, the model generalization (measured by the accuracy on unseen target domain) \lvv{may} not necessarily increase, but sometimes decreases instead, as the solid lines show.
On the one hand, this may be because the model does not best utilize the rich information of diversified domains; on the other hand, it may be due to the existence of low-quality samples which contain redundant or noisy information that is unprofitable to generalization~\cite{CleanNet}.
This discovery indicates that there is still room for improvement of the effectiveness of domain augmentation if we enable each domain to be certainly conducive to model generalization as the dash lines in Fig~\ref{fig:motivation}.

In this work, we therefore aim to ensure the strictly positive correlation between model generalization and domain diversity to guarantee and further enhance the effectiveness of domain augmentation.
To do this, we take inspiration from the literature of convex game that requires each player to bring profit to the coalition~\cite{ConvexGame,supermodularity,convexfuzzygame}, which is consistent to our key insight, i.e, make each domain bring benefit to model generalization. Thus, we propose to formalize DG as a convex game between domains.
First, we design a novel regularization term based on the supermodularity of convex game.
This regularization encourages each diversified domain to contribute to improving model generalization, thus enables the model to better exploit the diverse information.
In the meawhile, considering that there may exist samples with unprofitable or even harmful information to generalization, we further construct a sample filter based on the proposed regularization to get rid of the low-quality samples such as noisy or redundant ones, so that their deterioration to model generalization can be avoided.
We provide some heuristic analyses and intuitive explanations about the mechanisms behind to demonstrate the rationality in Section~\ref{sec:theory_proof}.

Nevertheless, it is well known that the supermodularity also indicates increasing marginal contribution, which may not hold intuitively in DG, where the marginal contribution of domains is generally decreasing. To mitigate the gap between theory and practice, we impose a constraint on the naive supermodularity when construct our regularization term. We constrain the regularization to work only in case that the supermodularity is violated, i.e., when the marginal contribution of domains decreases. Thus, the limit of our regularization optimization is actually to achieve a \emph{constant marginal contribution}, rather than an impracticable \emph{increasing marginal contribution}.
Hence, our regularization can additionally regularize the decreasing speed of the marginal contribution as slow as possible by optimizing towards the \emph{constant marginal contribution}, just like changing the line \textit{Ideal (a)} in Fig~\ref{fig:motivation} into line \textit{Ideal (b)}.
Generally, the role of our proposed supermodularity regularization is to encourage the contribution of each domain, and further relieve the \emph{decreasing marginal contribution} of domains to a certain extent, so as to better utilize the diversified information.

\textbf{Contributions.} Our contributions in this work include: (i) Exploring the relation of model generalization and source domain diversity, which reveals the limit of previous domain augmentation strand;
(ii) Introducing convex game into DG to guarantee and further enhance the validity of domain augmentation. The proposed framework encourages each domain to conducive to generalization while avoiding the negative impact of low-quality samples, enabling the model to better utilize the information within diversified domains;
(iii) Providing heuristic analysis and intuitive explanations about the rationality. The effectiveness and superiority are verified empirically across extensive real-world datasets.

\section{Related Work}

\textbf{Domain Generalization} researches out-of-distribution generalization with knowledge only extracted from multiple source domains.
A promising direction is to diversify training domains so as to improve generalization, referring as to domain augmentation~\cite{CrossGrad,AdvAug,MixStyle,FACT,L2A-OT}. L2A-OT~\cite{L2A-OT} creates pseudo-novel domains from source data by maximizing an optimal transport-based divergence measure. CrossGrad~\cite{CrossGrad} generates samples from fictitious domains via gradient-based domain perturbation while AdvAug~\cite{AdvAug} achieves so via adversarially perturbing images. 
MixStyle~\cite{MixStyle} and FACT~\cite{FACT} mix style information of different instances to synthetic novel domains.
Instead of enriching domain diversity, another popular solution that learning domain-invariant representations by distribution alignment via kernel-based optimization \cite{DICA,SCA}, adversarial learning \cite{MMD-AAE,CCSA}, \lv{or using uncertainty modeling~\cite{DSU}}
 demonstrate effectiveness for model generalization.
Other recent DG works also explore low-rank decomposition \cite{CSD}, self-supervised signals~\cite{Jigen}, gradient-guided dropout \cite{RSC}, etc. 
Though our proposed framework builds on the domain augmentation group, we aim to guarantee and further enhance their efficacy beyond via a convex game perspective.

\textbf{Convex Game} is a highly interesting class of cooperative games introduced by~\cite{ConvexGame}. A game is called convex when it satisfies the condition that the profit obtained by the cooperation of two coalitions plus the profit obtained by their intersection will not be less than the sum of profit obtained by the two respectively (a.k.a. supermodularity)~\cite{ConvexGame,supermodularity,convexfuzzygame}.
Co-Mixup\cite{Co-Mixup} 
formulates the optimal construction of mixup augmentation data while encouraging diversity among them by introducing supermodularity. Nevertheless, it is applied to supervised learning which aims to construct salience mixed samples.
Recently, ~\cite{onlineDG} rethinks the single-round minmax setting of DG and recasts it as a repeated online game between a player minimizing risk and an adversary presenting test distributions in light of online convex optimization~\cite{onlineconvexopti}. We note that the definition of convex game exploited in our work follows~\cite{ConvexGame}, distinct from that in~\cite{onlineDG, onlineconvexopti}.
To the best of our knowledge, this work is the first to introduce convex game into DG to enhance generalization capability.

\textbf{Meta Learning}~\cite{learning} is a long-term research exploring to learn how to train a particular model through the training of a meta-model~\cite{learntooptimize,MAML,fewshot}, and has drawn increasing attention from DG community~\cite{MetaReg,MASF,FCN,MLDG} recently. The main idea is to simulate domain shift during training by drawing virtual-train/test domains from the original source domains. 
MLDG~\cite{MLDG} originates the episode training paradigm from \cite{MAML}, back-propagating the second-order gradients from an ordinary task loss on random meta-test domains split from the source domains. 
Subsequent meta learning-based DG methods utilize a similar strategy to meta-learn a regularizer~\cite{MetaReg}, feature-critic network~\cite{FCN}, or semantic relationships~\cite{MASF}.
Different from the former paradigm that purely leverages the gradient of task objective, which may cause sub-optimal,
we utilize the ordinary task losses to construct a supermodularity regularization with more stable optimization, aiming to encourage each training domain to contribute to model generalization.

\section{Domain Convex Game}
\label{sec:method}
\begin{figure}[t]
  \centering
  \includegraphics[width=1.0\linewidth]{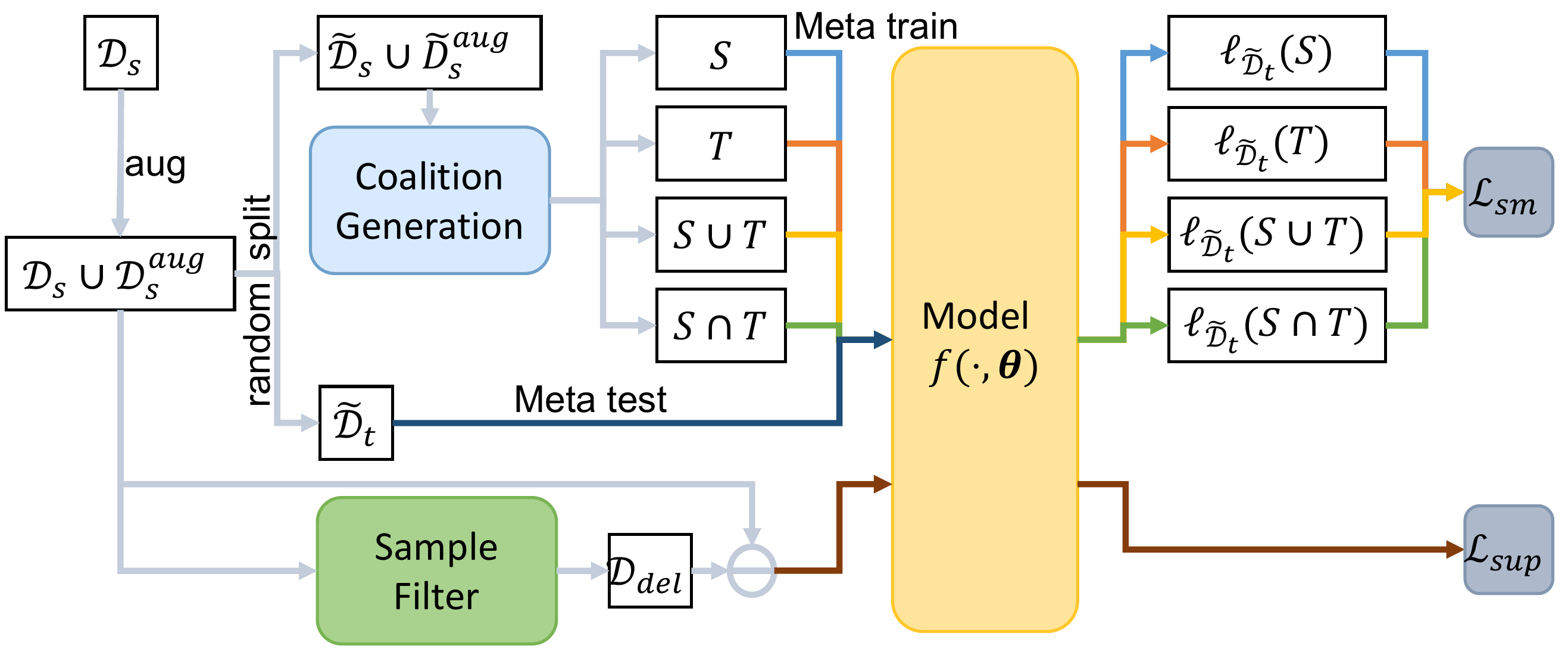}
  \caption{The pipeline of DCG. We first randomly split the diversified training domains into meta-train and meta-test domains, and generate four coalitions from the former according to the definition of convex game. Then we conduct meta learning on the four coalitions respectively and construct our regularization loss utilizing the meta-test losses of them based on the supermodularity. Meanwhile, we eliminate the low-quality samples by a sample filter and calculate supervision loss on the retained samples. }
  \label{fig:framework}
  \vspace{-4mm}
\end{figure}

Motivated by such an observation in Section~\ref{sec:intro}, we propose Domain Convex Game (DCG) framework to train models that can best utilize domain diversity, as illustrated in Fig.~\ref{fig:framework}.
First, we cast DG as a convex game between domains and design a novel regularization term employing the supermodularity, which encourages each domain to benefit model generalization. Further, we construct a sample filter based on the regularization to exclude bad samples that may cause negative effect on generalization. 
In this section, we define the problem setup and present the general form of DCG.

\subsection{Preliminary}
Assuming that there are $P$ source domains of data $\mathcal{D}_s = \cup_{k=1}^PD_k$ with $n_k$ labelled samples $\{(\boldsymbol{x}^k_i, y^k_i)\}_{i=1}^{n_k}$ in the $k$-th domain $D_k$, where $\boldsymbol{x}^k_i$ and $y^k_i \in \{1,2,\cdots,C\}$ denote the samples and corresponding labels. DG aims to train a domain-agnostic model $f(\cdot,\boldsymbol{\theta})$ parametrized with $\boldsymbol{\theta}$ on source domains that can generalize well on unseen target domain(s) $\mathcal{D}_t$.
As an effective solution for DG, domain augmentation aims to enrich the diversity of source domains generally by synthesizing novel domains via mixing domain-related information, hence boosting model generalization~\cite{L2A-OT,MixStyle,FACT}.
\lvv{Our work builds on this strand, and the key insight is to ensure and further improve its efficacy by better leveraging the domain diversity. For concision, in this paper, we adopt a simple Fourier-based augmentation technique~\cite{FACT,FDA} to prepare our diversified source domains. Note that the augmentation strategy is substitutable.}


\lvv{Technically}, owing to the property that the phase component of Fourier spectrum preserves high-level semantics of the original signal, while the amplitude component contains low-level statistics~\cite{1981fourier,1982fourier}, we augment the source data by distorting the amplitude information while keeping the phase information unchanged. Specifically, we mix the amplitude spectrum of an instance with that of another arbitrary instance by a linear interpolation strategy to synthesize augmented instances from novel domains. We refer readers to~\cite{FDA,FACT} for implementation details.
Since each augmented sample is generated by mixing domain information of sample pairs from random source domains in a random proportion, \lv{it has statistics distinct from the others so that can be regarded as drawn from a novel augmented domain}. Thus, we possess another $Q$ augmented source domains 
of data $\mathcal{D}_s^{aug} = \cup_{k=1}^QD_{P+k}$
with only one sample $\{(\boldsymbol{x}^{P+k}_i, y^{P+k}_i)\}_{i=1}^{1}$ in the $(P+k)$-th domain $D_{P+k}$, where $\boldsymbol{x}^{P+k}_i$ and $y^{P+k}_i$ denote the augmented samples and corresponding labels.
\lv{Note that the number of augmented domains generated this way is equivalent to the total number of all the original samples since each original sample pair will generate a pair of augmented samples.}
The goal of DCG is to train a generalizable model $f(\cdot,\boldsymbol{\theta})$ for unseen target domain(s) $\mathcal{D}_t$ with the aid of 
all $P+Q$ diversified source domains $\mathcal{D}_s\cup \mathcal{D}_s^{aug}$.

\subsection{Supermodularity Regularization Term}
\label{sec:sm_reg}
Let $M = \{1,2,\cdots,m\}$ be a finite set of players and $2^M$ is the family of $2^{|M|}$ subsets of $M$. A cooperative game with player set $M$ is a map $v:2^M \xrightarrow{} \mathbb{R}$. For coalition $S\in2^M$, $v(S)$ is called the worth of $S$, and is interpreted as the total profit that $S$ can obtain when the players in $S$ cooperate. 
A game is called convex if it satisfies the
\textit{supermodularity property}~\cite{convexfuzzygame,ConvexGame,supermodularity}, i.e., for each $S,T\in2^M$:
\begin{equation}
  v(S \cup T) + v(S \cap T) \ge v(S) + v(T).
  \label{eq:convex_sm}
\end{equation}
According to this definition, we can obtain: 
\begin{equation}
  v(S \cup \{i\} \cup \{j\}) - v(S \cup \{i\}) \ge v(S \cup \{j\}) - v(S),
\end{equation}
where $S\in2^M\verb|\|\{\varnothing\}$ and $i,j$ are two players not in $S$. 
We can see that convex game requires each player to contribute to the coalition, which is consistent with our key insight, that is, each training domain is expected to benefit model generalization. More than this, convex game also possesses \textit{increasing marginal contribution property} for players, which may not hold in DG. However, this property does not hinder our goal, but can further alleviate the \textit{decreasing marginal contribution} for domains, as discussed in Section~\ref{sec:intro}.

Thus, we first cast DG as a convex game between domains.
To achieve this, at each training iteration, we randomly split the original source data $\mathcal{D}_s$ into $P-V$ meta-train domains of data $\Tilde{\mathcal{D}_s}$ and $V$ meta-test domains of data $\tilde{\mathcal{D}}_t$, where $\Tilde{\mathcal{D}_s}$ and $\tilde{\mathcal{D}}_t$ share no domain. Then we pick out the augmented domains generated by data in $\Tilde{\mathcal{D}_s}$, denoted as $\tilde{\mathcal{D}}_s^{aug}$, and incorporate them into the meta-train domains. 
This strategy to conduct meta-train/test domains is to mimic the real train-test domain shift in domain augmentation strand, which is discussed in Section~\ref{sec:discussion_main}. 
Then, since one domain may contain multiple samples, we specifically consider involving a specific convex game: \emph{convex fuzzy game}~\cite{convexfuzzygame} where each player (i.e., each domain) can be partitioned into multiple parts (each part represents a sample in DG).
Now we have a finite set of partitioned players 
$\tilde{M} = \tilde{\mathcal{D}_s} \cup\tilde{\mathcal{D}}_s^{aug}$.
We can obtain coalitions $S,T \in 2^{\tilde{M}}$ by randomly sampling two sets of data from meta-train data $\tilde{\mathcal{D}_s} \cup\tilde{\mathcal{D}}_s^{aug}$, respectively. And $S\cup T, S\cap T$ can be naturally constructed by the union and intersection of $S$ and $T$.
As for the profit $v(O), O\in\{S,T,S\cup T, S\cap T\}$, we take the generalization performance evaluated on virtual-test domains $\tilde{\mathcal{D}_t}$ after the meta-training on each coalition $O$ as the value of profit $v(O)$.

Specifically, assuming a loss function $\ell(f(\boldsymbol{x},\boldsymbol{\theta}),y)$ for a sample between its output and label, e.g., cross-entropy loss for classification task, we first conduct virtual training on the four coalitions $\{S,T,S\cup T, S\cap T\}$, respectively, with the optimization objective:
\begin{equation}
  \small
  \mathcal{F}(O) := \sum_{x\in O}\ell(f(\boldsymbol{x},\boldsymbol{\theta}),y)), O \in \{S,T,S\cup T, S\cap T\}.
\end{equation}
Then the updated virtual parameters $\boldsymbol{\theta}'$ can be computed using one step of gradient descent:
\begin{equation}
  \boldsymbol{\theta}' = \boldsymbol{\theta} - \alpha \nabla_{\boldsymbol{\theta}}\mathcal{F}(O),
\end{equation}
where $\alpha$ is the virtual step size and is empirically set to be the same as the learning rate in our experiments. Thus, we can have the corresponding meta-test loss evaluated on the virtual-test domains $\tilde{\mathcal{D}_t}$ as below:
\begin{equation}
  \mathcal{G}(\boldsymbol{\theta}') := \mathbb{E}_{\boldsymbol{x} \in \tilde{\mathcal{D}_t}} \ell(f(\boldsymbol{x},\boldsymbol{\theta}'),y).
\end{equation}
This objective simulates test on unseen domains, thus can measure the model generalization obtained by training with one coalition, i.e., $v(O)=-\mathcal{G}(\boldsymbol{\theta}')$.
Hence, the supermodularity regularization can be constructed naturally utilizing the meta-test losses of the four coalitions based on Eq.~\eqref{eq:convex_sm}:
\begin{equation}
\small
\begin{split}
  \mathcal{L}_{sm} = \max\{0,&\mathcal{G}(\boldsymbol{\theta} - \alpha\nabla_{\boldsymbol{\theta}}\mathcal{F}(S\cup T)) + \mathcal{G}(\boldsymbol{\theta} - \alpha\nabla_{\boldsymbol{\theta}}\mathcal{F}(S\cap T))\\ - &\mathcal{G}(\boldsymbol{\theta} - \alpha\nabla_{\boldsymbol{\theta}}\mathcal{F}(S)) - \mathcal{G}(\boldsymbol{\theta} - \alpha\nabla_{\boldsymbol{\theta}}\mathcal{F}(T))\}.
\end{split}
\label{eq:l_reg}
\end{equation}
\lv{Here we exploit a $max(0,\cdot)$ function combined with the pure supermodularity to construct our regularization. In this way, $\mathcal{L}_{sm}>0$ only when the inequality in Eq.~\eqref{eq:convex_sm} is violated, i.e., the domain marginal contribution is decreasing. Thus, the limit of our regularization optimization corresponds to constant marginal contribution, not the inappropriate increasing marginal contribution.} Therefore, this regularization term can not only encourage each training domain to contribute to model generalization, but also alleviate the decrease of marginal contributions to some extent, enabling the model to fully leverage the rich information in diversified domains.

\subsection{Sample Filter}
Through the optimization of the regularization term, the model will be trained to better utilize the rich information of diversified source domains. However, what we cannot avoid is that there may exist some low-quality samples with harmful information to model generalization.
For instance, noisy samples will disturb model to learn generalizable knowledge; while redundant samples 
may lead to overfitting that hinder the model from learning more diverse patterns.

In this view, we further conduct a sample filter  to avoid the negative impact of low-quality samples.
Considering that the proposed regularization aims to penalize the decreasing marginal contribution of domains and then better utilize the diverse information, the samples that contribute more to the regularization loss (i.e., cause larger increase) are more unfavorable to our goal, hindering the improvement of model generalization.
Thus, we try to measure the contribution of each input to our regularization loss and define the contribution as its score.
Inspired by \cite{LRP} which defines the contribution of each input to the prediction by introducing layer-wise relevance propagation, we formulate the score of each input as the elementwise product between the input and its gradient to regularization loss, i.e., Input $\times$ Gradient:

\begin{equation}
  score = \boldsymbol{x}^T\nabla_{\boldsymbol{x}}\mathcal{L}_{sm},x\in \tilde{\mathcal{D}_s}\cup\tilde{\mathcal{D}}_s^{aug}.
  \label{eq:score}
\end{equation}
The higher the score of the sample, the greater the regularization loss will be increased caused by it, and the more it will hinder model from benefiting from diversified domains.
Therefore, we pick out the samples with the top-$k$ score, denoted as $\mathcal{D}_{del}$, and cast them away when calculating the supervision loss for diversified source domains \lv{to eliminate the negative effect of low quality samples}:
\begin{equation}
  \mathcal{L}_{sup} = \mathbb{E}_{\boldsymbol{x}\in\mathcal{D}_s\cup\mathcal{D}_s^{aug}\verb|\| \mathcal{D}_{del}} \ell(f(\boldsymbol{x},\boldsymbol{\theta}),y).
  \label{eq:l_cls}
\end{equation}
Thus, we optimize the regularization loss to enable model to better utilize the rich information within diversified domains. In the meanwhile, we eliminate the low-quality samples \lv{(e.g, noisy samples, redundant samples, etc)} by the sample filter to avoid their negative effects. \lv{Moreover, it is found that different types of low-quality samples are more likely to be discarded in different training stages, as discussed in Section~\ref{sec:theory_proof}. And we have explored out that low quality sample filtering is necessary for both original and augmented samples in Section~\ref{sec:discussion_main}.}

The overall optimization objective is:
\begin{equation}
  \arg \min_{\boldsymbol{\theta}} \mathcal{L}_{sup} + \omega \mathcal{L}_{sm},
  \label{eq:overall}
\end{equation}
where $\omega$ weights the supervision loss and the regularization term.
The overall methodological flow is illustrated schematically in Fig.~\ref{fig:framework} and summarized in Appendix~\ref{sec:alg}.
\section{Heuristic Analysis}
\label{sec:theory_proof}
In section~\ref{sec:method} we cast DG as a domain convex game and present the detailed formulation of our framework which revolves around the proposed regularization term.
Though this term is designed directly according to the supermodularity and has clear objective to achieve our goals, someone may still be curious about the mechanisms behind its effectiveness.
So in this section, we provide some heuristic analyses and intuitive explanations to further validate the rationality.

For brevity, we take $S=\{(\boldsymbol{x}_i, y_i)\}$ and $T=\{(\boldsymbol{x}_j,y_j)\}$ as an example, where $\boldsymbol{x}_i$ and $\boldsymbol{x}_j$ are from different domains.
According to Eq.~\eqref{eq:l_reg}, the optimization goal of our proposed regularization is to make the following inequality hold:
\begin{equation}\small
\begin{split}
    &\mathcal{G}(\boldsymbol{\theta} - \nabla_{\boldsymbol{\theta}}\ell(f(\boldsymbol{x}_i,\boldsymbol{\theta}),y_i) -\nabla_{\boldsymbol{\theta}}\ell(f(\boldsymbol{x}_j,\boldsymbol{\theta}),y_j))  + \mathcal{G}(\boldsymbol{\theta})\\
    & - \mathcal{G}(\boldsymbol{\theta} - \nabla_{\boldsymbol{\theta}} \ell(f(\boldsymbol{x}_i,\boldsymbol{\theta}),y_i))
     - \mathcal{G}(\boldsymbol{\theta} - \nabla_{\boldsymbol{\theta}} \ell(f(\boldsymbol{x}_j,\boldsymbol{\theta}),y_j)) \le 0.
\end{split}
\label{eq:example_obj}
\end{equation}
We then carry out the second-order Taylor expansion on the terms in Eq.~\eqref{eq:example_obj} and obtain:
\begin{equation}
\small
\begin{split}
    &(\nabla_i + \nabla_j)^T H (\nabla_i + \nabla_j) -  \nabla_i^T H \nabla_i - \nabla_j^T H \nabla_j \\
   & =\nabla_i^T H \nabla_j + \nabla_j^T H \nabla_i    \le 0   ,
\end{split}
\label{eq:taylor_expansion}
\end{equation}
$\nabla_i, \nabla_j$ denote $\nabla_{\boldsymbol{\theta}} \ell(f(\boldsymbol{x}_i,\boldsymbol{\theta}),y_i)$, $\nabla_{\boldsymbol{\theta}} \ell(f(\boldsymbol{x}_j,\boldsymbol{\theta}),y_j)$ respectively, $H = \frac{ \partial^{2} \mathcal{G}(\boldsymbol{\theta})}{\partial \boldsymbol{\theta} \partial \boldsymbol{\theta}^T}$ is the Hessian matrix of $\mathcal{G}(\boldsymbol{\theta})$.
We can see that all the zero- and first-order terms of the Taylor-expansion have been dissolved and only the second-order terms are left, which makes the optimization more stable.

Since Hessian matrix $H$ is a real symmetric matrix, for the case where $H$ is positive (negative) definite, we can perform Cholesky decomposition on $H (-H)$ as $L^T L$, where $L$ is an upper triangular matrix with real and positive diagonal elements. Thus, Eq.~\eqref{eq:taylor_expansion} can be further deduced as follows:
\begin{equation}
\small
\begin{split}
&\nabla_i^T H \nabla_j + \nabla_j^T H \nabla_i\\
& =
\begin{cases}
    (L \nabla_i)^T (L \nabla_j) + (L \nabla_j)^T (L \nabla_i) \le 0,& \text{$H \succ 0$,}\\
    -((L \nabla_i)^T (L \nabla_j) + (L \nabla_j)^T (L \nabla_i)) \le 0,& \text{$H \prec 0$.}
    \end{cases}
\end{split}
\label{eq:h_decomposition}
\end{equation}
Denote $L \nabla_i, L \nabla_j$ as $\tilde{\nabla_i}, \tilde{\nabla_j}$ respectively, which can be regarded as a mapping transformation of the original gradients. Specifically, $\nabla_i, \nabla_j$ are sample gradients generated in the original "training space" during the meta-training process, while $\tilde{\nabla_i}, \tilde{\nabla_j}$ are sample gradients transformed by matrix $L$. Since $L$ is derivated from the regularization term calculated on meta-test data that can indicate the model generalization, we can intuitively regard the transformed $\tilde{\nabla_i}, \tilde{\nabla_j}$ as sample gradients mapped to a "generalization space". Therefore, constraining sample gradients in this mapped "generalization space" may generalize better on the real test set compared to constraining the gradients in the original "training space".

Then two main cases can be analysed respectively.
\begin{case}
For Hessian matrix $H \prec 0$ (a.k.a. negative definite), Eq.~\eqref{eq:taylor_expansion} holds when \lv{${\tilde{\nabla_i}}^T\tilde{\nabla_j} \ge 0$}.
\label{cond:non-positive}
\end{case}

\textbf{\textit{mechanism.}}
When $H \prec 0$, i.e., achieving local maxima, which suggests inferior model generalization, the proposed regularization would help the model improve by enforcing domain consistency on discriminability, that is, pulling the samples from different classes apart and bringing the ones from the same class closer \lv{in the "generalization space"}. 
As for sample filtering, samples that possess inconsistent gradients, e.g., noisy samples, are more prone to be discarded.

\textbf{\textit{analysis.}}
\lv{As Eq.~\eqref{eq:h_decomposition} shows, our regularization aims to make the inner product of transformed sample gradients positive when $H \prec 0$, i.e., make the sample gradients consistent in the "generalization space".}
Assuming samples $\boldsymbol{x}_i$, $\boldsymbol{x}_j$ belong to the same class, then their transformed gradients will be inconsistent when they are apart in the "generalization space", and be consistent when they are close.  
In contrast, if $\boldsymbol{x}_i$, $\boldsymbol{x}_j$ are from different classes, their transformed gradients would certainly be inconsistent if the samples are close, since they share the same model while possessing different labels. Thus, the optimization of our regularization will draw the samples from the same class closer while pulling the ones from different classes apart to make the gradients consistent, which enforces domain consistency on discriminability. 
\lv{As for sample filtering, the samples that possess very inconsistent gradients are contrary to our goal most and are more likely to obtain larger scores, which are generally noise samples since they are often located at outliers. Therefore, the noise samples are more prone to be discarded in this case.}

\begin{case}
For Hessian matrix $H \succ 0$ (a.k.a. positive definite), Eq.~\eqref{eq:taylor_expansion} holds when \lv{${\tilde{\nabla_i}}^T\tilde{\nabla_j} \le 0$}.
\label{cond:positive}
\end{case}

\textbf{\textit{mechanism.}}
When $H \succ 0$, i.e., achieving local optima, the proposed regularization would help the model jump out by further squeezing out the information within hard samples, that is, detecting the hard samples and then assigning them larger weights implicitly. As for sample filtering, samples that possess very consistent gradients, e.g., redundant samples, are more prone to be discarded.

\textbf{\textit{analysis.}}
As Eq.~\eqref{eq:h_decomposition} shows, our regularization aims to make the inner product of the transformed sample gradients negative  when $H \succ 0$, i.e., make the gradients inconsistent in the "generalization space". This objective is contrary to our main supervision loss that aims to make all the samples clustered, so it can be regarded as an adversarial optimization. Concretely, the regularization enables the model to generate and detect samples with inconsistent gradients which generally be hard samples since they are often far away from the class center. Then these hard samples would contribute more to the main supervision loss and thus can be considered as being assigned larger weights implicitly during the optimization, just like the mechanism of focal loss~\cite{focal_loss}. Thus, our regularization can help model jump out of the local optima by squeezing out more information within hard samples, avoiding the model depending on easy patterns or even overfitting on redundant ones. 
For sample filtering, the samples that produce very consistent gradients, which also means they are redundant ones to a certain, are more likely to be detrimental to our regularization loss and be filtered.

For the general case that $H$ is not fully positive or negative definite, we can take SVD decomposition and regard the model as combined by positive or negative definite sub-matrices. Then our conclusion holds for each subspace represented by each submatrix.

\section{Experiments}
\label{sec:exp}

\subsection{Dataset and Implementation Details}
\label{sec:dataset}
To evaluate our method, we conduct extensive experiments on three popular benchmarks for DG:
\noindent\textbf{PACS}\cite{pacs} is an object recognition benchmark that covers 9991 images of 7 categories from four different domains, i.e., Art, Cartoon, Photo and Sketch, which with large discrepancy in image styles.
\noindent\textbf{Office-Home}\cite{home} is a commonly-used benchmark including four domains (Art, Clipart, Product, RealWorld). It contains
15,500 images of 65 classes in total.
\noindent\textbf{mini-DomainNet}\cite{dael} is a very large-scale domain generalization benchmark consists of about 140k images with 126 classes from four different domains (Clipart, Painting, Real, Sketch).
For all benchmarks, we conduct the commonly used leave-one-domain-out experiments~\cite{DBA} and adopt ResNet-18/50 pre-trained on ImageNet~\cite{resnet} as backbone.
We train the network using mini-batch SGD with batch size 16, momentum 0.9 and weight decay 5e-4.
The initial learning rate is 0.001 and decayed by 0.1 at 80\% of the total epochs. 
For hyper-parameters, we set $\omega = 0.1$ and $k=5$ for all experiments, which are selected on validation set following standard protocol.
All results are reported based on the average accuracy over three independent runs. More details and results with error bars are provided in Appendix.

\subsection{Experimental Results}

\begin{table}
  \centering
  \small
      \resizebox{\columnwidth}{!}{
      \begin{tabular}{l|cccc|c}
      \toprule
      Methods & Art & Cartoon & Photo & Sketch & Avg. \\
      \midrule
      \multicolumn{6}{c}{\textit{ResNet18}} \\
      \midrule
      DeepAll\cite{FACT} & 77.63 & 76.77 & 95.85 & 69.50 & 79.94 \\
      MLDG~\cite{MLDG} & 78.70 & 73.30 & 94.00 & 65.10 & 80.70 \\
      MASF~\cite{MASF} & 80.29 & 77.17 & 94.99 & 71.69 & 81.04 \\
      L2A-OT~\cite{L2A-OT} & 83.30 & 78.20 & \underline{96.20} & 73.60 & 82.80 \\
      DDAIG~\cite{DEEPALL} & 84.20 & 78.10 & 95.30 & 74.70 & 83.10 \\
      RSC~\cite{RSC} & 83.43 & \underline{80.31} & 95.99 & 80.85 & 85.15 \\
      MixStyle ~\cite{MixStyle} & 84.10 &  78.80 & 96.10 &  75.90 & 83.70 \\
      FACT \cite{FACT}& \underline{85.37} &78.38& 95.15& 79.15 &84.51 \\
      \lv{DSU~\cite{DSU} } & 83.60 &79.60& 95.80& 77.60& 84.10\\
      \lvv{STNP}~\cite{STNP} & 84.41 & 79.25 & 94.93 & \textbf{83.27} & \underline{85.47}\\
      \midrule
      DCG (\textit{ours})  & \textbf{85.94} &	\textbf{80.76}	& \textbf{96.41} &	\underline{82.08}	& \textbf{86.30} \\
      \midrule
      \multicolumn{6}{c}{\lvv{\textit{ResNet50}}} \\
      \midrule
      DeepAll~\cite{FACT} & 84.94 & 76.98 & 97.64 & 76.75 & 84.08 \\
      RSC~\cite{RSC} & 87.89 & 82.16 & \underline{97.92} & 83.35 & 87.83 \\
      FACT \cite{FACT} & 89.63 & 81.77 & 96.75 & 84.46 & 88.15 \\
      DDG~\cite{DDG}  &88.90& \underline{85.00}& 97.20& 84.30& 88.90\\
      PCL~\cite{PCL} &90.20& 83.90& \textbf{98.10}& 82.60& 88.70\\
      STNP~\cite{STNP} &\textbf{90.35}& 84.20& 96.73& \underline{85.18}& \underline{89.11}\\
      \midrule
      DCG (\textit{ours}) & \underline{90.24} & \textbf{85.12} & 97.76 & \textbf{86.31} & \textbf{89.84} \\
      \bottomrule
      \end{tabular}}
      \vspace{-2mm}
      \caption{Leave-one-domain-out results on PACS.}
    \label{tab:pacs_res18}
  \vspace{-3mm}
\end{table}

\noindent\textbf{Results on PACS} \lvv{based on ReNet-18 and ResNet-50 are summarized in Table~\ref{tab:pacs_res18}. 
It is clear that DCG achieves the best performance among all the competitors \lvv{on both backbones}.
We notice that DCG surpasses the Fourier based augmentation method FACT by a large margin of $1.8\%$ and $1.7\%$ on ResNet-18 and ResNet-50, respectively, which indicate the importance of encouraging each domain to contribute to model generalization.
Especially, on the harder target domains Cartoon and Sketch, our method still outperforms the SOTA. 
There also exist cases where DCG performs relatively poorly, this may due to the task is relatively simple (e.g. \textit{photo}).
In general, the comparisons reveal the effectiveness of DCG and further demonstrate that the convex game between domains improves model generalization.}

\begin{table}
    \centering
    \small
        \resizebox{\columnwidth}{!}{
        \begin{tabular}{l|cccc|c}
        \toprule
        Methods & Art & Clipart & Product & Real & Avg. \\
        \midrule
        DeepAll  & 57.88 & 52.72 & 73.50 & 74.80 & 64.72 \\
        MLDG~\cite{MLDG} &52.88 & 45.72 & 69.90 & 72.68 & 60.30 \\
        SagNet~\cite{SagNet}  & 60.20  &45.38& 70.42& 73.38& 62.34\\
        RSC~\cite{RSC}   & 58.42 & 47.90 & 71.63 & 74.54 & 63.12 \\
        DDAIG~\cite{DEEPALL} & 59.20 & 52.30 & 74.60 & 76.00 & 65.50 \\
        L2A-OT~\cite{L2A-OT} & \underline{60.60} & 50.10 & \underline{74.80} & \textbf{77.00} & 65.60 \\
        MixStyle~\cite{MixStyle} & 58.70 & 53.40 & 74.20 & 75.90 & 65.50\\
        FACT \cite{FACT}& 60.34 & 54.85 & 74.48 & 76.55 & \underline{66.56} \\
        \lv{DSU~\cite{DSU}} & 60.20 & 54.80 & 74.10 & 75.10 & 66.10 \\
        \lvv{STNP~\cite{STNP}} & 59.55 & \underline{55.01} & 73.57 & 75.52 & 65.89\\
        \midrule
        DCG (\textit{ours}) & \textbf{60.67} &	\textbf{55.46} &	\textbf{75.26}	& \underline{76.82} &	\textbf{67.05}  \\
        \bottomrule
        \end{tabular}}
        \vspace{-2mm}
        \caption{Leave-one-domain-out results on Office-Home.
        }
      \label{tab:officehome}
      \vspace{-5mm}
  \end{table}

  \noindent\textbf{Results on Office-Home} \lvv{based on ReNet-18} are presented in Table~\ref{tab:officehome}, where we beat all the compared baselines in terms of the average accuracy. 
  Due to the similarity to the pre-trained dataset ImageNet,
  DeepAll acts as a strong baseline on Office-Home.
  Many previous DG methods, such as MLDG, SagNet, and RSC, can not improve over the simple DeepAll baseline.
  Nevertheless, our DCG achieves a consistent improvement over DeepAll on all the held-out domains. 
  Moreover, DCG surpasses the latest domain augmentation methods L2A-OT and FACT. The incremental advantages may be due to the relatively smaller domain shift, where the decreasing marginal contribution of domains is more severe. 

\begin{table}
  \centering
  \small
    \resizebox{\columnwidth}{!}{
    \begin{tabular}{l|cccc|c}
    \toprule
    Methods & Clipart & Painting & Real & Sketch & Avg. \\
    \midrule
    DeepAll & 65.30 &	58.40	&64.70&	59.00&	61.86  \\
    ERM~\cite{ERM}  & 65.50 & 57.10 & 62.30 & 57.10 & 60.50 \\
     MLDG~\cite{MLDG} & 65.70 & 57.00 & 63.70 & 58.10 & 61.12 \\
     Mixup~\cite{mixup} & \underline{67.10} & 59.10 & 64.30 & 59.20 & 62.42 \\
    MMD~\cite{MMD} & 65.00 & 58.00 & 63.80 & 58.40& 61.30 \\
    SagNet~\cite{SagNet} & 65.00 & 58.10 & 64.20 & 58.10 & 61.35 \\
     CORAL~\cite{CORAL} & 66.50 & \underline{59.50} & \underline{66.00} & \underline{59.50} & \underline{62.87} \\
     MTL~\cite{MTL} & 65.30 & 59.00 & 65.60 & 58.50 & 62.10 \\
    \midrule
     DCG (\textit{ours}) & \textbf{69.38} &	\textbf{61.79}&	\textbf{66.34}&	\textbf{63.21}&	\textbf{65.18}  \\
    \bottomrule
    \end{tabular}}
    \vspace{-2mm}
    \caption{Leave-one-domain-out results on mini-DomainNet. 
    }
  \label{tab:domainnet}
  \vspace{-5mm}
\end{table}

\begin{figure*}[t]
  \centering
  \vspace{-1mm}
   \includegraphics[width=1.0\linewidth]{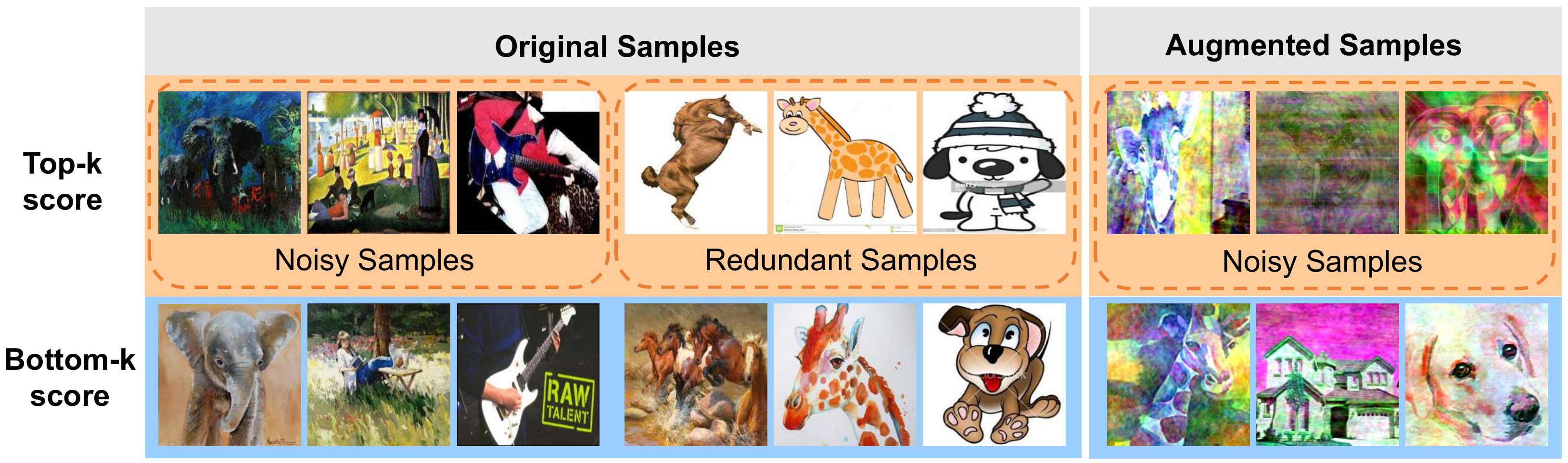}
  \vspace{-7mm}
   \caption{The visualization of samples with top-$k$ and bottom-$k$ score respectively with Cartoon as the unseen target domain. }
   \label{fig:visualization}
   \vspace{-3mm}
\end{figure*}

\noindent\textbf{Results on Mini-DomainNet} \lvv{based on ReNet-18} are shown in Table~\ref{tab:domainnet}. The much larger number of categories and images makes DomainNet a much more challenging benchmark. DCG still achieves the state-of-the-art performance of $65.18\%$, surpassing the SOTA by a large margin of $2.31\%$. \lvv{It indicates that the waste of diversified information in large datasets is more serious, further validating our efficacy.}

\subsection{Analysis}

\noindent\textbf{Ablation Study.}
In Table~\ref{tab:ablation}, we investigate the role of each component in DCG, including Fourier augmentation (Aug.), supermodularity regularization (Reg. ($\mathcal{L}_{sm}$)) and sample filter (Filter. ($\mathcal{F}_{sm}$)).
The Baseline is trained only with the supervision loss of all the original source data. We incorporating our supermodularity regularization $\mathcal{L}_{sm}$ with the Fourier augmentation to obtain Model 3, which greatly surpasses Model 1, demonstrating the significance of encouraging each diversified domain to contribute to generalization. 
Besides, we aslo apply a regularization $\mathcal{L}_{maml}$ which sums the meta-test losses of all the tasks as MAML~\cite{MAML} to conduct Model 2, its inferiority to Model 3 indicates conducting 
\begin{table}
  \setlength{\tabcolsep}{0.4mm}
  \centering
  \small
  \centering
    \resizebox{\columnwidth}{!}{
    \begin{tabular}{c|ccc|cccc|c}
    \toprule
    Method & Aug. & Reg. & Filter. & Art & Cartoon & Photo & Sketch & Avg.\\
    \midrule
    Baseline & - & - & - & 77.6 & 76.8 & 95.9 & 69.5 & \lv{79.9}\\
    \midrule
    Model 1 & $\checkmark$ & - & - & 83.9& 77.0 & 95.6 & 77.4 & \lv{83.4}\\
    Model 2 & $\checkmark$ & $\mathcal{L}_{maml}$ & - & 84.7 &	79.0 &	95.7 &	80.1 &	\lv{84.9}	\\
    Model 3 & $\checkmark$ & $\mathcal{L}_{sm}$ & - &85.1 &	80.1 &	95.9 &	81.4 & \lv{85.6} \\
    Model 4 & $\checkmark$ & - & $\mathcal{F}_{maml}$ & 84.1 &	77.7 &	95.5 &	78.2 &	\lv{83.9} \\
    Model 5 & $\checkmark$ & - & $\mathcal{F}_{sm}$ & 84.4 &	78.2 &	95.8 &	79.3 &	\lv{84.4} \\
    Model 6 & $\checkmark$ & $\mathcal{L}_{maml}$ & $\mathcal{F}_{maml}$ & 85.3 & 79.9 & 96.0 &	81.5 &	\lv{85.7} \\
    \midrule
    DCG & $\checkmark$ & $\mathcal{L}_{sm}$ & $\mathcal{F}_{sm}$ & \textbf{85.9} &	\textbf{80.8}&	\textbf{96.4} &	\textbf{82.1} &	\textbf{86.3}\\
    \bottomrule
    \end{tabular}}
    \vspace{-2mm}
    \caption{Ablation study of DCG on PACS dataset.}
  \label{tab:ablation}
 \vspace{-5mm}
\end{table}
convex game between domains is more helpful to generalization than simply applying the meta loss. 
Comparing Model 5 with Model 1, we can observe that the proposed sample filter is also conducive to generalization, suggesting the importance of eliminating nonprofitable information. Finally, DCG performs best in all variants, indicating that the two proposed components complement and benefit each other.

\begin{figure*}
  \centering
  \begin{subfigure}{0.23\linewidth}
  \includegraphics[width=1.0\linewidth]{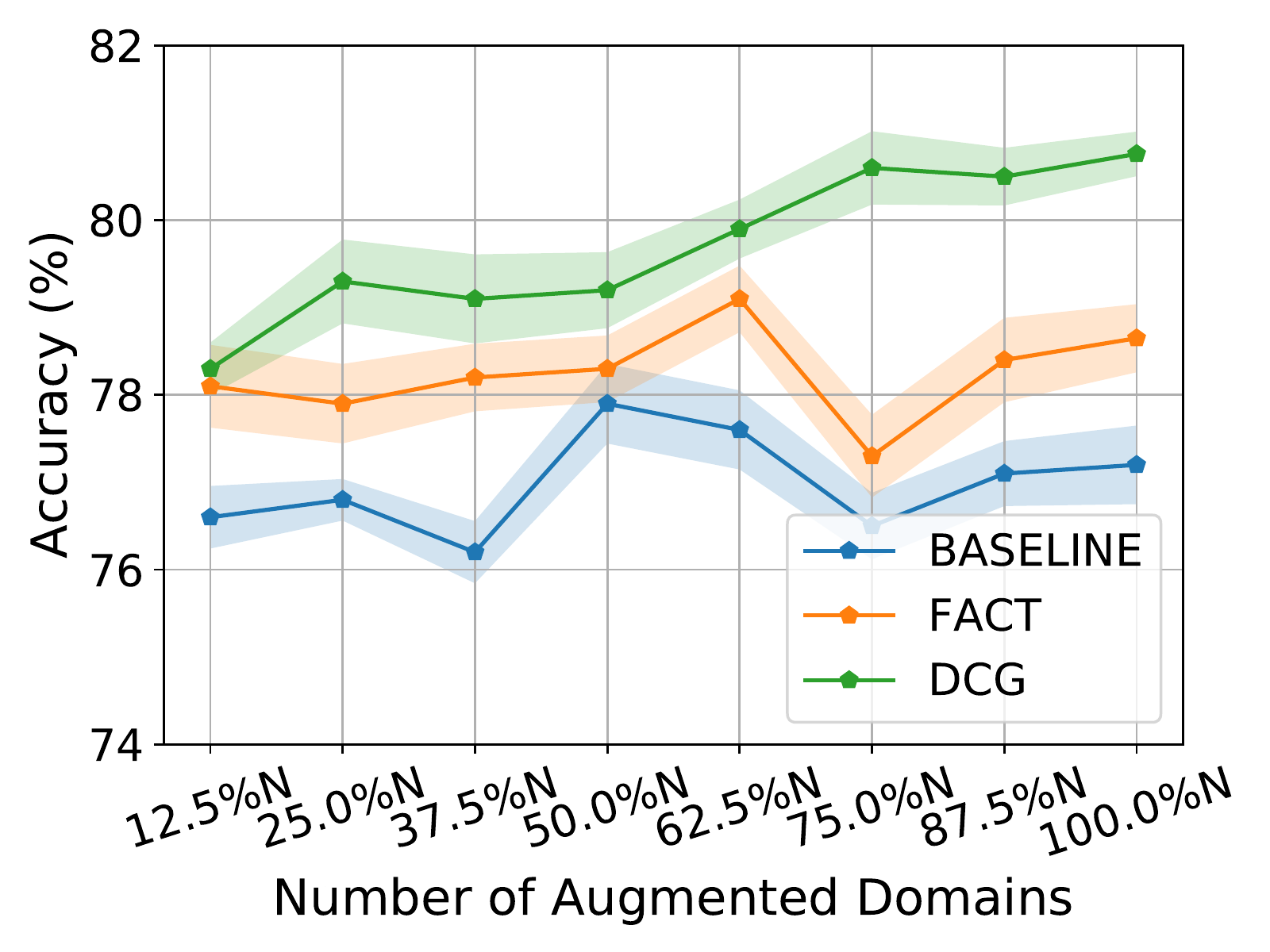}
    \caption{Cartoon.}
    \label{fig:inc_cartoon}
  \end{subfigure}
  \hfill
  \begin{subfigure}{0.23\linewidth}
  \includegraphics[width=1.0\linewidth]{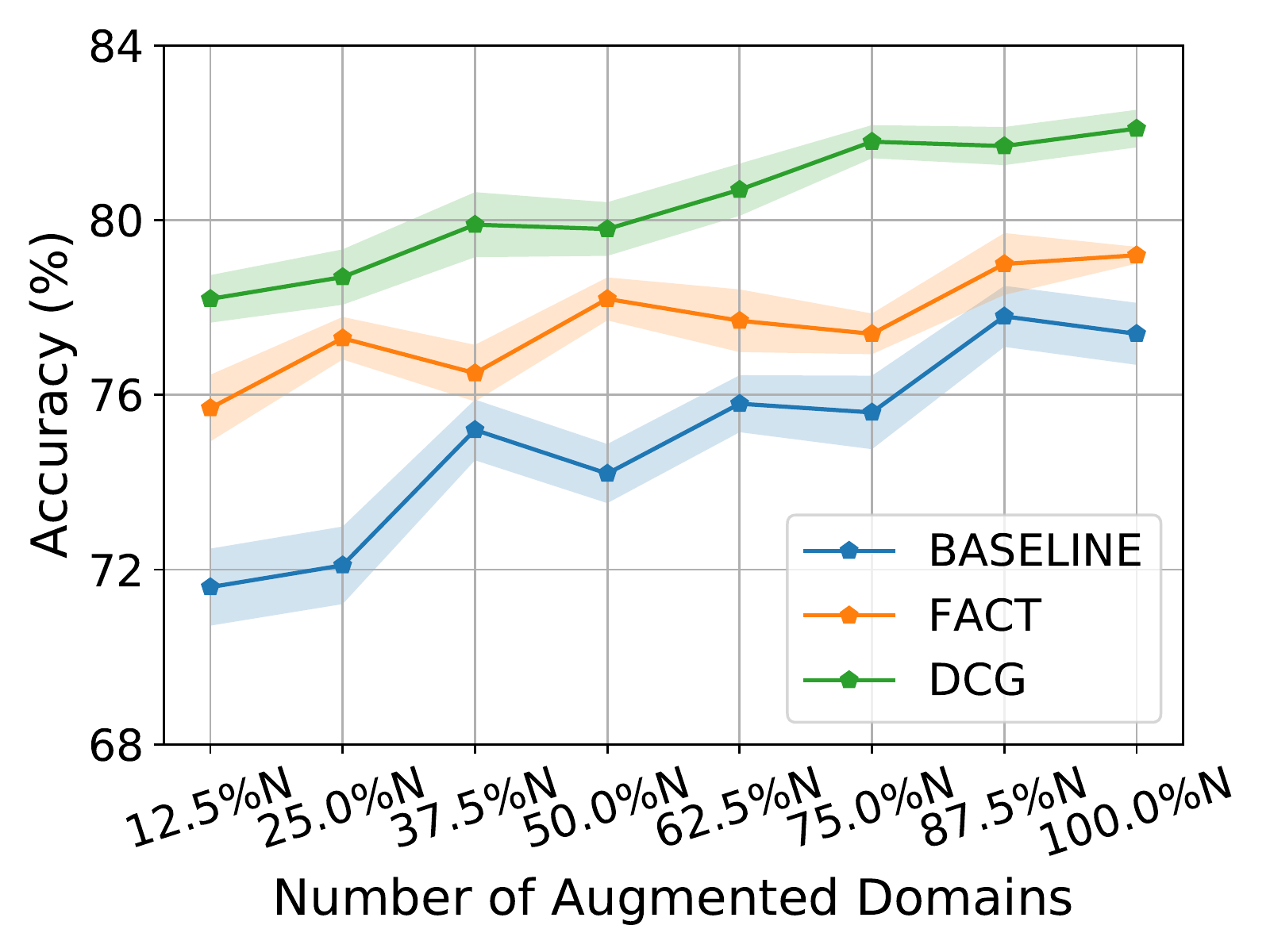}
    \caption{Sketch.}
    \label{fig:inc_sketch}
  \end{subfigure}
  \hfill
  \begin{subfigure}{0.23\linewidth}
    \includegraphics[width=1.0\linewidth]{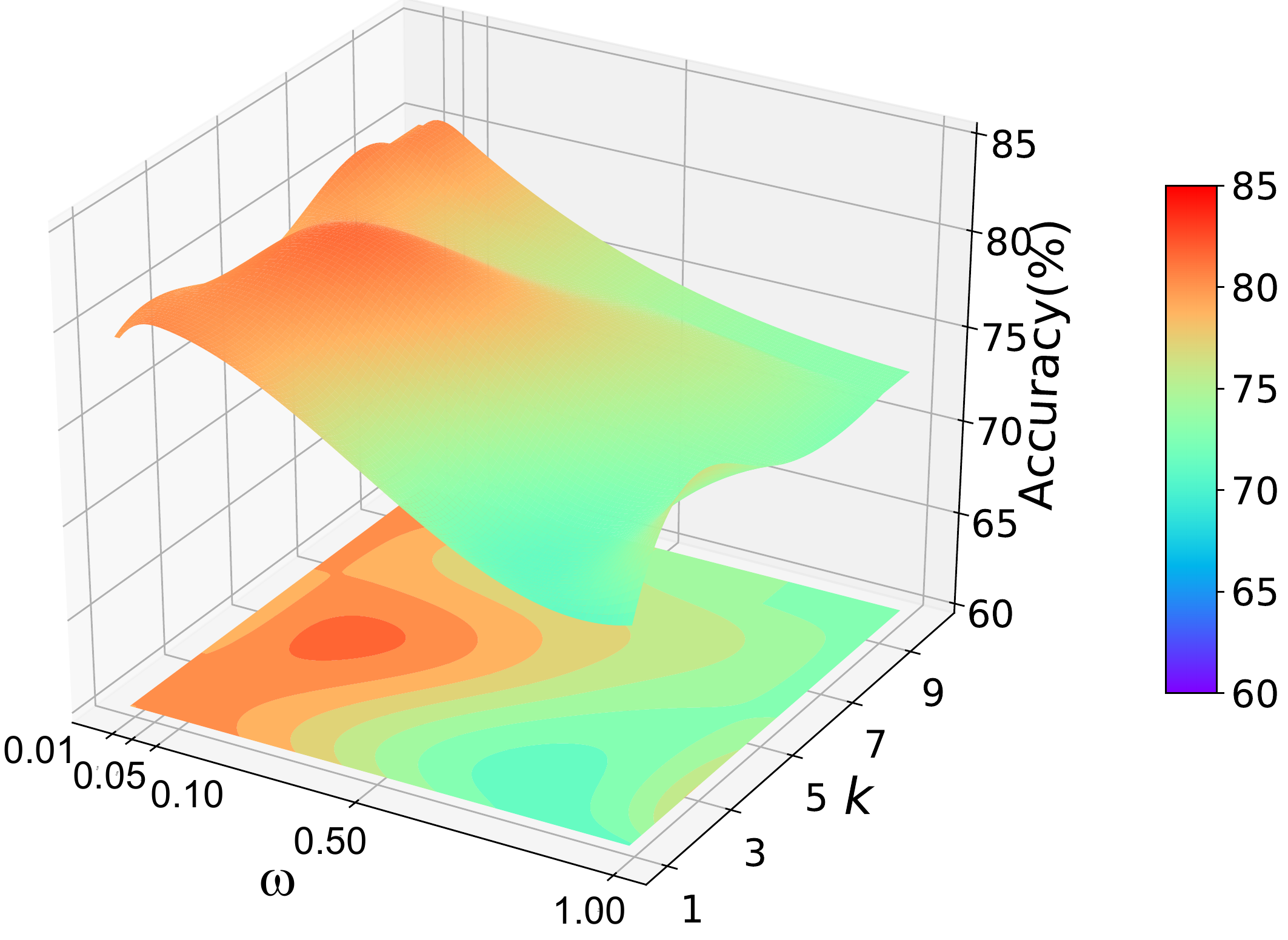}
      \caption{Cartoon.}
      \label{fig:sens_cartoon}
    \end{subfigure}
    \hfill
    \begin{subfigure}{0.23\linewidth}
    \includegraphics[width=1.0\linewidth]{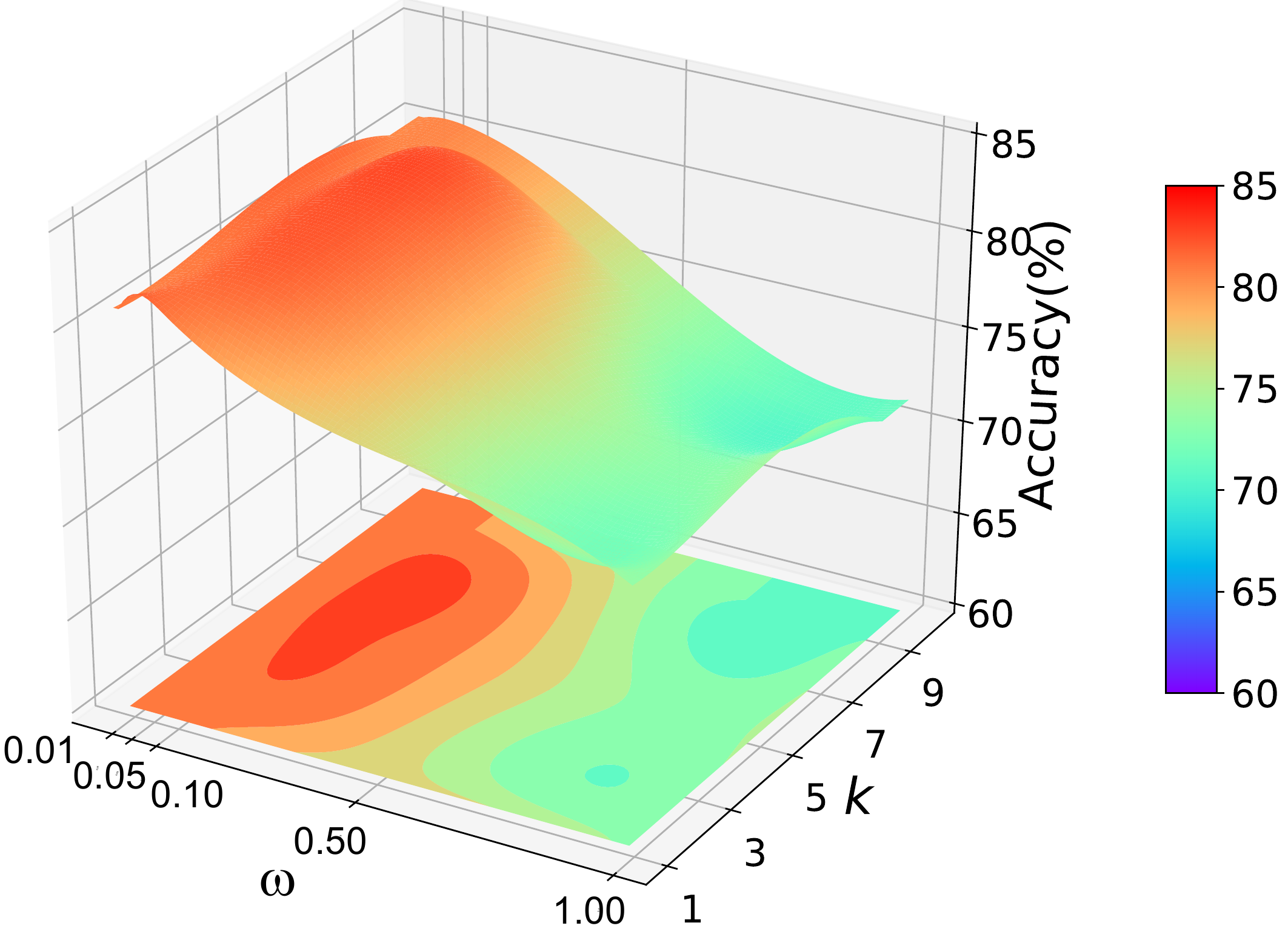}
      \caption{Sketch.}
      \label{fig:sens_sketch}
    \end{subfigure}
    \vspace{-2mm}
  \caption{(a)(b): relation between model generalization 
and domain diversity; (c)(d): sensitivity to hyper-parameters $\omega$ and $k$; with Cartoon and Sketch on PACS
dataset as the unseen target domain.}
  \vspace{-5mm}
\end{figure*}

\noindent\textbf{Generalization with Domain Diversity.}
Figure~\ref{fig:inc_cartoon},~\ref{fig:inc_sketch} show the model generalization with the increase of domain diversity. We use the classification accuracy on the held-out target domain as the metric of model generalization across domains, and the number of augmented domains to measure the domain diversity.
It is clear that on both Cartoon and Sketch tasks, the model generalization capability of the baseline methods do not necessarily improve with the increase of domain diversity, but sometimes decrease instead. While in our DCG, the model generalization increases monotonically with the domain diversity on the whole and the decrease of marginal contribution of domains is alleviated. 
Meanwhile, in a few cases, the generalization of DCG drops a little when domain diversity increases. This is reasonable since the additional augmented domains may be low-quality or harmful to generalization. 
The results demonstrate that our framework indeed encourages each diversified domain to contribute to model generalization, hence guarantee and further improve the performance of domain augmentation methods.

\noindent\textbf{Visualization of Filtered Samples.}
To visually verify that our sample filter can effectively eliminate low-quality samples, we provide the samples that obtain the top-$k$ / bottom-$k$ score the most times in the whole training process in Figure~\ref{fig:visualization}.
We can see that the discarded original samples with top-$k$ score in the first row either be noisy images that have messy background and fuzzy objects, or be images containing naive or classical information which may be redundant. While the high-quality original images in the bottom row are all vivid and rich in information. As for the augmented samples, the discarded ones are almost distinguishable while the retained high-quality ones are limpid.
These comparisons demonstrate the effectiveness of our sample filter.

\noindent\textbf{Sensitivity of Hyper-parameters.}
Figure~\ref{fig:sens_cartoon},~\ref{fig:sens_sketch} show the sensitivity of DCG to hyper-parameters $\omega$ and $k$. Specifically, the value of $\omega$ varies from $\{0.01, 0.05, 0.1, 0.5, 1,0\}$, while
$k$ changes from $\{1, 3, 5, 7, 9\}$.  
It can be observed
that DCG achieves competitive performances robustly under a wide range of hyper-parameter values, i.e., $0.05 \le \omega \le 0.3$ and $3 \le k \le 7$, in either task Cartoon or Sketch, which further verifies the stability of our method.

\subsection{Discussion}
\label{sec:discussion_main}

\begin{table}
  \centering
  \small
 \resizebox{\columnwidth}{!}{
 \begin{tabular}{c|cccc|c}
    \toprule
     Methods & Art & Cartoon & Photo & Sketch & Avg.\\
    \midrule
     Random\_meta\_split &85.6	&80.2&	96.0&	81.8&	85.9\\
    \midrule
     Filter\_only\_on\_aug & 85.4 & 80.6 & 96.7 &	81.8 &	86.1 \\
     Filter\_only\_on\_ori & 85.2 &80.0 & 96.5 &	82.3 &	86.0\\
    \midrule
    DCG & \textbf{85.9} &	\textbf{80.8}&	\textbf{96.4} &	\textbf{82.1} &	\textbf{86.3}\\
    \bottomrule
    \end{tabular}}
    \vspace{-2mm}
    \caption{Leave-one-domain-out results on PACS.}
    \label{tab:discussion}
    \vspace{-5mm}
\end{table}
\noindent\textbf{How to conduct the meta-train and meta-test domains?}
In DCG, we considers all the diversified domains $\mathcal{D}_s \cup \mathcal{D}_s^{aug}$ into training. We first randomly split the original source domains $\mathcal{D}_s$ into meta-train and meta-test domains, next pick out the domains in $\mathcal{D}_s^{aug}$ that are augmented by the current meta-train domains and then merge them into together. Thus, there is no domain augmented by the meta-test domains in the meta-train domains, and vice versa. However, why don't we also randomly split $\mathcal{D}_s^{aug}$ into two parts, since each diversified domain can be regarded as a novel domain?
We conduct experiments of this variant and the results in Table~\ref{tab:discussion} shows inferior performance to DCG. 
This may be because the synthetic novel domains still contain part of the domain-related information of the original ones. In this view, the strategy to conduct meta-train/test domains in Section~\ref{sec:sm_reg} can guarantee the meta-test domains are completely unseen, which better simulates the domain shift between diversified source domains and the held-out unseen target domain.

\noindent\textbf{Is low-quality sample filtering necessary for both original and augmented samples?}
We conduct experiments that apply the proposed sample filter only on the original samples or augmented samples and the results are shown in Table~\ref{tab:discussion}. 
It can be seen that both variants suffer from a performance drop, which indicates that there exist low-quality samples among both original and augmented samples.
Limiting the filtering range will make some low-quality samples be retained to participate in the training process, which may damage the model generalization.
Besides, the performance of only filtering the original samples is slightly lower than that of only filtering the augmented ones, which should be due to the augmented samples being less natural.

\vspace{2mm}
\section{Conclusion \lv{\& Limitation}}
\label{sec:conclusion}
\vspace{1mm}
\lvv{This work explores the relation of model generalization and domain diversity, aiming to guarantee and further enhance the efficacy of domain augmentation strand. We then propose a framework to enable each diversified domain contribute to generalization by casting DG as 
a convex game between domains.
Heuristic analysis and comprehensive experiments demonstrate our rationality and effectiveness.}
Note that we mainly focus on the mixup-based domain augmentation techniques for clarity, while the extension of DCG to other GAN-based techniques needs to be further explored. Besides, it also remains an open problem to design a more efficient strategy to avoid the decrease in training efficiency caused by meta-learning. 
Nevertheless, we believe our work can inspire the future work of enriching domain diversity 
with improved generalization capability.

\lvv{\paragraph{Acknowledgements.} This paper was supported by National Key R$\&$D Program of China (No. 2021YFB3301503), and also supported by the National Natural Science Foundation of China under Grant No. 61902028.}

\clearpage
{\small
\bibliographystyle{ieee_fullname}
\bibliography{egbib}

\begin{thebibliography}{10}\itemsep=-1pt

\bibitem{LRP}
Sebastian Bach, Alexander Binder, Grégoire Montavon, Frederick Klauschen,
  Klaus-Robert Müller, and Wojciech Samek.
\newblock On pixel-wise explanations for non-linear classifier decisions by
  layer-wise relevance propagation.
\newblock {\em PLoS ONE}, 10(7), 2015.

\bibitem{MetaReg}
Yogesh Balaji, Swami Sankaranarayanan, and Rama Chellappa.
\newblock Metareg: Towards domain generalization using meta-regularization.
\newblock In {\em NeurIPS}, pages 1006--1016, 2018.

\bibitem{MTL}
Gilles Blanchard, Aniket~Anand Deshmukh, {\"{U}}r{\"{u}}n Dogan, Gyemin Lee,
  and Clayton Scott.
\newblock Domain generalization by marginal transfer learning.
\newblock {\em J. Mach. Learn. Res.}, pages 2:1--2:55, 2021.

\bibitem{convexfuzzygame}
Rodica Br{\^a}nzei, Dinko Dimitrov, and Stef Tijs.
\newblock Convex fuzzy games and participation monotonic allocation schemes.
\newblock {\em Fuzzy sets and systems}, 139(2):267--281, 2003.

\bibitem{Jigen}
Fabio~Maria Carlucci, Antonio D'Innocente, Silvia Bucci, Barbara Caputo, and
  Tatiana Tommasi.
\newblock Domain generalization by solving jigsaw puzzles.
\newblock In {\em CVPR}, pages 2229--2238, 2019.

\bibitem{MASF}
Qi Dou, Daniel~Coelho de Castro, Konstantinos Kamnitsas, and Ben Glocker.
\newblock Domain generalization via model-agnostic learning of semantic
  features.
\newblock In {\em NeurIPS}, pages 6447--6458, 2019.

\bibitem{MAML}
Chelsea Finn, Pieter Abbeel, and Sergey Levine.
\newblock Model-agnostic meta-learning for fast adaptation of deep networks.
\newblock In {\em ICML}, volume~70, pages 1126--1135, 2017.

\bibitem{SCA}
Muhammad Ghifary, David Balduzzi, W.~Bastiaan Kleijn, and Mengjie Zhang.
\newblock Scatter component analysis: {A} unified framework for domain
  adaptation and domain generalization.
\newblock {\em TPAMI}, 39(7):1414--1430, 2017.

\bibitem{DeepLearning}
Ian~J. Goodfellow, Yoshua Bengio, and Aaron~C. Courville.
\newblock {\em Deep Learning}.
\newblock Adaptive computation and machine learning. {MIT} Press, 2016.

\bibitem{onlineconvexopti}
Elad Hazan.
\newblock Introduction to online convex optimization.
\newblock {\em Found. Trends Optim.}, 2(3-4):157--325, 2016.

\bibitem{resnet}
Kaiming He, Xiangyu Zhang, Shaoqing Ren, and Jian Sun.
\newblock Deep residual learning for image recognition.
\newblock In {\em CVPR}, pages 770--778, 2016.

\bibitem{RSC}
Zeyi Huang, Haohan Wang, Eric~P. Xing, and Dong Huang.
\newblock Self-challenging improves cross-domain generalization.
\newblock In {\em ECCV}, pages 124--140, 2020.

\bibitem{supermodularity}
Tatsuro Ichiishi.
\newblock Super-modularity: applications to convex games and to the greedy
  algorithm for lp.
\newblock {\em Journal of Economic Theory}, 25(2):283--286, 1981.

\bibitem{STNP}
Juwon Kang, Sohyun Lee, Namyup Kim, and Suha Kwak.
\newblock Style neophile: Constantly seeking novel styles for domain
  generalization.
\newblock In {\em CVPR}, pages 7130--7140, June 2022.

\bibitem{Co-Mixup}
Jang{-}Hyun Kim, Wonho Choo, Hosan Jeong, and Hyun~Oh Song.
\newblock Co-mixup: Saliency guided joint mixup with supermodular diversity.
\newblock In {\em ICLR}, 2021.

\bibitem{DeepLearning2}
Yann LeCun, Yoshua Bengio, and Geoffrey~E. Hinton.
\newblock Deep learning.
\newblock {\em Nat.}, 521(7553):436--444, 2015.

\bibitem{MMD}
Chen{-}Yu Lee, Tanmay Batra, Mohammad~Haris Baig, and Daniel Ulbricht.
\newblock Sliced wasserstein discrepancy for unsupervised domain adaptation.
\newblock In {\em CVPR}, pages 10285--10295, 2019.

\bibitem{CleanNet}
Kuang{-}Huei Lee, Xiaodong He, Lei Zhang, and Linjun Yang.
\newblock Cleannet: Transfer learning for scalable image classifier training
  with label noise.
\newblock In {\em CVPR}, pages 5447--5456, 2018.

\bibitem{DBA}
Da Li, Yongxin Yang, Yi{-}Zhe Song, and Timothy~M. Hospedales.
\newblock Deeper, broader and artier domain generalization.
\newblock In {\em ICCV}, pages 5543--5551, 2017.

\bibitem{MLDG}
Da Li, Yongxin Yang, Yi{-}Zhe Song, and Timothy~M. Hospedales.
\newblock Learning to generalize: Meta-learning for domain generalization.
\newblock In {\em AAAI}, pages 3490--3497, 2018.

\bibitem{pacs}
Da Li, Yongxin Yang, Yi-Zhe Song, and Timothy~M Hospedales.
\newblock Deeper, broader and artier domain generalization.
\newblock In {\em ICCV}, pages 5542--5550, 2017.

\bibitem{MMD-AAE}
Haoliang Li, Sinno~Jialin Pan, Shiqi Wang, and Alex~C. Kot.
\newblock Domain generalization with adversarial feature learning.
\newblock In {\em CVPR}, pages 5400--5409, 2018.

\bibitem{learntooptimize}
Ke Li and Jitendra Malik.
\newblock Learning to optimize.
\newblock In {\em ICLR}, 2017.

\bibitem{DSU}
Xiaotong Li, Yongxing Dai, Yixiao Ge, Jun Liu, Ying Shan, and Lingyu Duan.
\newblock Uncertainty modeling for out-of-distribution generalization.
\newblock In {\em ICLR}, 2022.

\bibitem{FCN}
Yiying Li, Yongxin Yang, Wei Zhou, and Timothy~M. Hospedales.
\newblock Feature-critic networks for heterogeneous domain generalization.
\newblock In {\em ICML}, pages 3915--3924, 2019.

\bibitem{focal_loss}
Tsung{-}Yi Lin, Priya Goyal, Ross~B. Girshick, Kaiming He, and Piotr
  Doll{\'{a}}r.
\newblock Focal loss for dense object detection.
\newblock In {\em ICCV}, pages 2999--3007. {IEEE} Computer Society, 2017.

\bibitem{RTN}
Mingsheng Long, Han Zhu, Jianmin Wang, and Michael~I. Jordan.
\newblock Unsupervised domain adaptation with residual transfer networks.
\newblock In {\em NeurIPS}, pages 136--144, 2016.

\bibitem{MMAN}
Xinhong Ma, Tianzhu Zhang, and Changsheng Xu.
\newblock Deep multi-modality adversarial networks for unsupervised domain
  adaptation.
\newblock {\em {IEEE} Trans. Multim.}, 21(9):2419--2431, 2019.

\bibitem{CCSA}
Saeid Motiian, Marco Piccirilli, Donald~A. Adjeroh, and Gianfranco Doretto.
\newblock Unified deep supervised domain adaptation and generalization.
\newblock In {\em ICCV}, pages 5716--5726, 2017.

\bibitem{DICA}
Krikamol Muandet, David Balduzzi, and Bernhard Sch{\"{o}}lkopf.
\newblock Domain generalization via invariant feature representation.
\newblock In {\em ICML}, pages 10--18, 2013.

\bibitem{SagNet}
Hyeonseob Nam, HyunJae Lee, Jongchan Park, Wonjun Yoon, and Donggeun Yoo.
\newblock Reducing domain gap via style-agnostic networks.
\newblock {\em CoRR}, abs/1910.11645, 2019.

\bibitem{1981fourier}
A.~V. Oppenheim and J.~S. Lim.
\newblock The importance of phase in signals.
\newblock {\em Proc IEEE}, 69(5):529--541, 1981.

\bibitem{TransferLearning}
Sinno~Jialin Pan and Qiang Yang.
\newblock A survey on transfer learning.
\newblock {\em {IEEE} Trans. Knowl. Data Eng.}, 22(10):1345--1359, 2010.

\bibitem{CORAL}
Xingchao Peng, Qinxun Bai, Xide Xia, Zijun Huang, Kate Saenko, and Bo Wang.
\newblock Moment matching for multi-source domain adaptation.
\newblock In {\em ICCV}, pages 1406--1415, 2019.

\bibitem{1982fourier}
L.~N. Piotrowski and F.~W. Campbell.
\newblock A demonstration of the visual importance and flexibility of
  spatial-frequency amplitude and phase.
\newblock {\em Perception}, 11(3):337--46, 1982.

\bibitem{CSD}
Vihari Piratla, Praneeth Netrapalli, and Sunita Sarawagi.
\newblock Efficient domain generalization via common-specific low-rank
  decomposition.
\newblock In {\em ICML}, pages 7728--7738, 2020.

\bibitem{fewshot}
Sachin Ravi and Hugo Larochelle.
\newblock Optimization as a model for few-shot learning.
\newblock In {\em ICLR}, 2017.

\bibitem{onlineDG}
Elan Rosenfeld, Pradeep Ravikumar, and Andrej Risteski.
\newblock An online learning approach to interpolation and extrapolation in
  domain generalization.
\newblock {\em CoRR}, abs/2102.13128, 2021.

\bibitem{CrossGrad}
Shiv Shankar, Vihari Piratla, Soumen Chakrabarti, Siddhartha Chaudhuri, Preethi
  Jyothi, and Sunita Sarawagi.
\newblock Generalizing across domains via cross-gradient training.
\newblock In {\em ICLR}, 2018.

\bibitem{ConvexGame}
Lloyd~S Shapley.
\newblock Cores of convex games.
\newblock {\em International journal of game theory}, 1(1):11--26, 1971.

\bibitem{DistantTL}
Ben Tan, Yu Zhang, Sinno~Jialin Pan, and Qiang Yang.
\newblock Distant domain transfer learning.
\newblock In {\em AAAI}, pages 2604--2610, 2017.

\bibitem{robustness}
Rohan Taori, Achal Dave, Vaishaal Shankar, Nicholas Carlini, Benjamin Recht,
  and Ludwig Schmidt.
\newblock Measuring robustness to natural distribution shifts in image
  classification.
\newblock In {\em NeurIPS}, 2020.

\bibitem{learning}
Sebastian Thrun and Lorien Pratt.
\newblock {\em Learning to learn}.
\newblock Springer Science \& Business Media, 2012.

\bibitem{ERM}
Vladimir Vapnik.
\newblock An overview of statistical learning theory.
\newblock {\em {IEEE} Trans. Neural Networks}, 10(5):988--999, 1999.

\bibitem{home}
Hemanth Venkateswara, Jose Eusebio, Shayok Chakraborty, and Sethuraman
  Panchanathan.
\newblock Deep hashing network for unsupervised domain adaptation.
\newblock In {\em CVPR}, pages 5018--5027, 2017.

\bibitem{AdvAug}
Riccardo Volpi, Hongseok Namkoong, Ozan Sener, John~C. Duchi, Vittorio Murino,
  and Silvio Savarese.
\newblock Generalizing to unseen domains via adversarial data augmentation.
\newblock In {\em NeurIPS}, pages 5339--5349, 2018.

\bibitem{DomainAug}
Keyulu Xu, Mozhi Zhang, Jingling Li, Simon~Shaolei Du, and Stefanie Jegelka.
\newblock How neural networks extrapolate: From feedforward to graph neural
  networks.
\newblock In {\em ICLR}, 2021.

\bibitem{FACT}
Qinwei Xu, Ruipeng Zhang, Ya Zhang, Yanfeng Wang, and Qi Tian.
\newblock A fourier-based framework for domain generalization.
\newblock In {\em CVPR}, pages 14383--14392, 2021.

\bibitem{FDA}
Yanchao Yang and Stefano Soatto.
\newblock {FDA:} fourier domain adaptation for semantic segmentation.
\newblock In {\em CVPR}, pages 4084--4094, 2020.

\bibitem{PCL}
Xufeng Yao, Yang Bai, Xinyun Zhang, Yuechen Zhang, Qi Sun, Ran Chen, Ruiyu Li,
  and Bei Yu.
\newblock {PCL:} proxy-based contrastive learning for domain generalization.
\newblock In {\em CVPR}, pages 7087--7097, 2022.

\bibitem{mixup}
Hongyi Zhang, Moustapha Ciss{\'{e}}, Yann~N. Dauphin, and David Lopez{-}Paz.
\newblock mixup: Beyond empirical risk minimization.
\newblock In {\em ICLR}, 2018.

\bibitem{DDG}
Hanlin Zhang, Yi{-}Fan Zhang, Weiyang Liu, Adrian Weller, Bernhard
  Sch{\"{o}}lkopf, and Eric~P. Xing.
\newblock Towards principled disentanglement for domain generalization.
\newblock In {\em CVPR}, pages 8014--8024, 2022.

\bibitem{DEEPALL}
Kaiyang Zhou, Yongxin Yang, Timothy~M. Hospedales, and Tao Xiang.
\newblock Deep domain-adversarial image generation for domain generalisation.
\newblock In {\em AAAI}, pages 13025--13032, 2020.

\bibitem{L2A-OT}
Kaiyang Zhou, Yongxin Yang, Timothy~M. Hospedales, and Tao Xiang.
\newblock Learning to generate novel domains for domain generalization.
\newblock In {\em ECCV}, pages 561--578, 2020.

\bibitem{dael}
Kaiyang Zhou, Yongxin Yang, Yu Qiao, and Tao Xiang.
\newblock Domain adaptive ensemble learning.
\newblock {\em IEEE Transactions on Image Processing}, 30:8008--8018, 2021.

\bibitem{MixStyle}
Kaiyang Zhou, Yongxin Yang, Yu Qiao, and Tao Xiang.
\newblock Domain generalization with mixstyle.
\newblock In {\em ICLR}, 2021.

\end{thebibliography}
}

\clearpage
\appendix

  \section{Social Impact}
  \label{sec:impact}
  Our work focuses on domain generalization and attempts to make each training domain contribute to model generalization, which validates and further enhances the effectiveness of domain augmentation strand.
  This method produces a positive impact on the society and  community, saves the cost and time of data annotation, boosts the reusability of knowledge across domains, and greatly improves the efficiency. Nevertheless, this work suffers from some negative influences, which is worthy of further research and exploration. Specifically, more jobs of classification or target detection for rare or variable conditions may be cancelled. Moerover, we should be cautious about the result of the failure of the system, which could render people believe that classification was unbiased. Still, it might be not, which might be misleading, e.g., when using the system in a highly variable unseen target domain.
  
  \section{Algorithm of DCG}
  \label{sec:alg}
  
  \lvv{In this work, we propose a Domain Convex Game (DCG) framework to guarantee and further
  enhance the validity of domain augmentation approaches by casting DG as a convex game between domains. Here, we summarize the training process of DCG based on the discussions in main body as Algorithm~\ref{alg:DCG}.}

  \begin{algorithm}[H]
    \vspace{-1mm}
    \caption{The Algorithm of Domain Convex Game.}
    \label{alg:DCG}
    \begin{algorithmic} [1]
    \REQUIRE $P+Q$ diversified source domains $\mathcal{D}_s \cup \mathcal{D}_s^{aug}$; Hyper-parameters: $\omega, k$.
     \STATE randomly initialize model parameters $\boldsymbol{\theta}$.
    \FOR{iter in iterations}
    \STATE Randomly sample a mini-batch of $\mathcal{D}_s$ as $B$ and a mini-batch of $\mathcal{D}_s^{aug}$ as $B^{aug}$. 
    \STATE Split: $\tilde{\mathcal{D}_s}$ and $\tilde{\mathcal{D}_t}$ $\xleftarrow{} B$, Pick out: $\tilde{\mathcal{D}}_s^{aug}$ from $B^{aug}$.
    \STATE Construct coalitions $S,T$ by randomly sampling from $\tilde{\mathcal{D}_s} \cup \tilde{\mathcal{D}}_s^{aug}$; construct coalitions $S\cup T, S\cap T$.
    \STATE Calculate supermodularity regularization loss $\mathcal{L}_{sm}$ as Eq.~\eqref{eq:l_reg}.
    \STATE Pick out low-quality samples $\mathcal{D}_{del}$ with the top-$k$ score calculated by Eq.~\eqref{eq:score}.
    \STATE Calculate supervision loss $\mathcal{L}_{sup}$ as Eq.~\eqref{eq:l_cls}.
    \STATE Update $\boldsymbol{\theta} = \arg \min_{\boldsymbol{\theta}} \mathcal{L}_{sup} + \omega\mathcal{L}_{sm}$.
    \ENDFOR
    \end{algorithmic}
  \end{algorithm}

  \section{Experimental Details}
  \label{sec:implementation}
  For all benchmarks, we conduct the commonly used leave-one-domain-out experiments~\cite{DBA}, where we choose one domain as the unseen target domain for evaluation, and train the model on all remaining domains. 
  We adopt the standard augmentation protocol as in~\cite{Jigen}, all images are resized to 224 × 224, following with random  resized cropping, horizontal flipping and color jittering. And the Fourier domain augmentation strategy utilized to diversify source domains closely follows the implementations in~\cite{FACT}. 
  \lvv{The network backbone is set to ResNet-18 or ResNet-50 pre-trained on ImageNet~\cite{resnet} following other related works.} 
  We train the network using mini-batch SGD with batch size 16, momentum 0.9 and weight decay 5e-4 for 50 epochs. The initial learning rate is 0.001 and decayed by 0.1 at 80\% of the total epochs. The meta step size $\alpha$ is set to be the same as the learning rate.
  For the hyper-parameters, i.e., the weight of regularization loss $\omega$ and the number of discarded bad samples in each iteration $k$, their values are selected on validation data following standard practice, where we use 90\% of available data as training data and 10\% as validation data. Specifically, we set $\omega = 0.1$ and $k=5$ for all experiments.
  Our framework is implemented with PyTorch on NVIDIA GeForce RTX 3090 GPUs. All results are reported based on the average accuracy over three independent runs for a fair comparison.

  \section{Additional Results}
  \lvv{\subsection{Time cost analysis}}
  \lvv{We conduct experiments to study the efficiency of our method in the training and inference stages respectively, and the results are shown in Table~\ref{tab:time_cost}. For the training stage, the time cost of DCG is indeed relatively high, which is due to
 the use of meta learning when constructing the regularization term and the backpropagation when calculating the score for sample filter. For substitute, we may edit the backpropagation path that computes gradients of inputs only on a smaller subnetwork to reduce time cost. Besides, we can see that for the inference stage, our DCG method is as efficient as other methods and does not incur additional time costs.  Note that this work is an innovative effort to study the relation between model generalization and domain diversity, which is in a preliminary stage. And we will further explore more efficient techniques in future research.}
  
 \begin{table}[htbp]
  \centering
    \setlength{\tabcolsep}{1.0mm}{
    \begin{tabular}{l|cc}
    \toprule
    Methods & Training & Inference\\
    \midrule
    DEEPALL~\cite{FACT} & 168 s & 5 s\\
    FACT~\cite{FACT} & 186 s & 5 s\\
    MLDG~\cite{MLDG} & 275 s & 5 s\\
    \midrule
    DCG w$/$o Filter. & 349 s & 5 s\\
    DCG & 467 s & 5 s\\
    \bottomrule
    \end{tabular}}
    \caption{Running Time per Epoch.}
  \label{tab:time_cost}
\end{table}

\begin{table*}[htpb]
  \centering
    \setlength{\tabcolsep}{4.0mm}{
    \begin{tabular}{l|cccc|c}
    \toprule
    Methods & Art & Cartoon & Photo & Sketch & Avg. \\
    \midrule
    \multicolumn{6}{c}{\textit{ResNet18}} \\
    \midrule\
    MLDG~\cite{MLDG} & 78.70 & 73.30 & 94.00 & 65.10 & 80.70 \\
    L2A-OT~\cite{L2A-OT} & 83.30 & 78.20 & 96.20 & 73.60 & 82.80 \\
    RSC~\cite{RSC} & 83.43 & 80.31 & 95.99 & 80.85 & 85.15 \\
      DSU \cite{DSU} & 83.60 & 79.60 & 95.80 & 77.60 & 84.10\\
    \midrule
    DeepAll\cite{DEEPALL} & 77.63$\pm$0.84 & 76.77$\pm$0.33 & 95.85$\pm$0.20 & 69.50$\pm$1.26 & 79.94 \\
     MASF~\cite{MASF} & 80.29$\pm$0.18& 77.17$\pm$0.08& 94.99$\pm$0.09& 71.69$\pm$0.22& 81.04 \\
    DDAIG~\cite{DEEPALL} & 84.20$\pm$0.30& 78.10$\pm$0.60& 95.30$\pm$0.40& 74.70$\pm$0.80 & 83.10 \\
    MixStyle ~\cite{MixStyle} & 84.10$\pm$0.40 &  78.80$\pm$0.40 & 96.10$\pm$0.30 &  75.90$\pm$0.90 & 83.70 \\
    FACT \cite{FACT}& 85.37$\pm$0.29 &78.38$\pm$0.29& 95.15$\pm$0.26& 79.15$\pm$0.69 &	84.51 \\
    STNP \cite{STNP} & 84.41$\pm$0.62 & 79.25$\pm$0.98 & 94.93$\pm$0.07 & \textbf{83.27$\pm$2.03} & 85.47\\
    \midrule
    DCG (\textit{ours})  & \textbf{85.94$\pm$0.21} &	\textbf{80.76$\pm$0.36}	& \textbf{96.41$\pm$0.17} &	82.08$\pm$0.44	& \textbf{86.30} \\
  \midrule
  \multicolumn{6}{c}{\textit{ResNet50}} \\
  \midrule
  RSC~\cite{RSC} & 87.89 & 82.16 & 97.92 & 83.35 & 87.83 \\
  PCL \cite{PCL}& 90.20 &83.90& \textbf{98.10}& 82.60 &	88.70 \\
  \midrule
  DeepAll\cite{DEEPALL} & 84.94$\pm$0.66 & 76.98$\pm$1.13 & 97.64$\pm$0.10 & 76.75$\pm$0.41 & 84.08 \\
  FACT \cite{FACT}& 89.63$\pm$0.51 &81.77$\pm$0.19& 96.75$\pm$0.10& 84.46$\pm$0.78 &	88.15 \\
  DDG \cite{DDG}& 88.90$\pm$0.60 &85.00$\pm$1.90& 97.20$\pm$1.20& 84.30$\pm$0.70 &	88.90 \\
  STNP \cite{STNP} & \textbf{90.35$\pm$0.62} & 84.20$\pm$1.43 & 96.73$\pm$0.46 & 85.18$\pm$0.46 & 89.11\\
  \midrule
  DCG (\textit{ours})  & 90.24$\pm$0.48 &	\textbf{85.12$\pm$0.79}	& 97.76$\pm$0.13 &	\textbf{86.31$\pm$0.64}	& \textbf{89.84} \\
 
    \bottomrule
    \end{tabular}}
    \caption{Leave-one-domain-out results on PACS.}
  \label{tab:pacs_all}
\end{table*}

\begin{table*}[htpb]
  \centering
   \setlength{\tabcolsep}{4.0mm}{
    \begin{tabular}{l|cccc|c}
    \toprule
    Methods & Art & Clipart & Product & Real & Avg. \\
    \midrule
    MLDG~\cite{MLDG} &52.88 & 45.72 & 69.90 & 72.68 & 60.30 \\
    SagNet \cite{SagNet} & 60.20 & 45.38 & 70.42 & 73.38 & 62.34\\
    RSC~\cite{RSC}   & 58.42 & 47.90 & 71.63 & 74.54 & 63.12 \\
    L2A-OT~\cite{L2A-OT} & 60.60 & 50.10 & 74.80 & \textbf{77.00} & 65.60 \\
    DSU \cite{DSU} & 60.20 & 54.80 & 74.10 & 75.10 & 66.10\\
    \midrule
    DeepAll \cite{DEEPALL} & 57.88$\pm$0.20 & 52.72$\pm$0.50& 73.50$\pm$0.30& 74.80$\pm$0.10 & 64.72 \\
    DDAIG~\cite{DEEPALL} & 59.20$\pm$0.10& 52.30$\pm$0.30& 74.60$\pm$0.30& 76.00$\pm$0.10 & 65.50 \\
    MixStyle \cite{MixStyle} & 58.70$\pm$0.30 & 53.40$\pm$0.20 & 74.20$\pm$0.10 & 75.90$\pm$0.10 & 65.50\\
    FACT \cite{FACT}& 60.34$\pm$0.11& 54.85$\pm$0.37& 74.48$\pm$0.13& 76.55$\pm$0.10 & 66.56 \\
    STNP \cite{STNP} & 59.55$\pm$0.21 & 55.01$\pm$0.29 & 73.57$\pm$0.28 & 75.52$\pm$0.21 & 65.89\\
    \midrule
     DCG (\textit{ours}) & \textbf{60.67$\pm$0.14} &	\textbf{55.46$\pm$0.32} &	\textbf{75.26$\pm$0.18}	& 76.82$\pm$0.09 &	\textbf{67.05}  \\
    \bottomrule
    \end{tabular}}
    \caption{Leave-one-domain-out results on Office-Home.
    }
  \label{tab:home_all}
\end{table*}

\subsection{Experimental Results with Error Bars}
\label{sec:additional_res} 
For the sake of objective, we run all the experiments multiple times with random seed. We report the average results in the main body of paper for elegant, and show the complete results with error bars in the form of mean$\pm$std below (Table.~\ref{tab:pacs_all},~\ref{tab:home_all},~\ref{tab:domainnet_all}).

\begin{table*}[b]
  \centering
   \setlength{\tabcolsep}{4.0mm}{
    \begin{tabular}{l|cccc|c}
    \toprule
    Methods & Clipart & Painting & Real & Sketch & Avg. \\
    \midrule
    DeepAll \cite{DEEPALL} & 65.30 &	58.40	&64.70&	59.00&	61.86  \\
    \midrule
    ERM~\cite{ERM}  & 65.50 $\pm$ 0.3& 57.10 $\pm$ 0.5& 62.30 $\pm$ 0.2& 57.10 $\pm$ 0.1 & 60.50 \\
     MLDG~\cite{MLDG} & 65.70 $\pm$ 0.2& 57.00 $\pm$ 0.2& 63.70 $\pm$ 0.3& 58.10 $\pm$ 0.1 & 61.12 \\
     Mixup~\cite{mixup} & 67.10 $\pm$ 0.2& 59.10 $\pm$ 0.5& 64.30 $\pm$ 0.3& 59.20 $\pm$ 0.3 & 62.42 \\
    MMD~\cite{MMD} & 65.00 $\pm$ 0.5& 58.00 $\pm$ 0.2& 63.80 $\pm$ 0.2& 58.40 $\pm$ 0.7& 61.30 \\
    SagNet~\cite{SagNet} & 65.00 $\pm$ 0.4& 58.10 $\pm$ 0.2 &64.20 $\pm$ 0.3& 58.10 $\pm$ 0.4 & 61.35 \\
     CORAL~\cite{CORAL} & 66.50 $\pm$ 0.2& 59.50 $\pm$ 0.4 &66.00 $\pm$ 0.6& 59.50 $\pm$ 0.1 & 62.87 \\
     MTL~\cite{MTL} & 65.30 $\pm$ 0.5 &59.00 $\pm$ 0.4& 65.60 $\pm$ 0.4& 58.50 $\pm$ 0.2 & 62.10 \\
    \midrule
     DCG (\textit{ours}) & \textbf{69.38$\pm$0.19} &	\textbf{61.79$\pm$0.22}&	\textbf{66.34$\pm$0.27}&	\textbf{63.21$\pm$0.09}&	\textbf{65.18}  \\
    \bottomrule
    \end{tabular}}
    \caption{Leave-one-domain-out results on Mini-DomainNet.
}
  \label{tab:domainnet_all}
\end{table*}

\end{document}